\documentclass[11pt]{article}

\usepackage{acl}

\usepackage{times}
\usepackage{latexsym}
\usepackage[T1]{fontenc}
\usepackage[utf8]{inputenc}
\usepackage{microtype}
\usepackage{inconsolata}

\usepackage{graphicx}
\usepackage{booktabs}
\usepackage{multirow}
\usepackage{amsmath}
\usepackage{amssymb}
\usepackage{xcolor}
\usepackage{hyperref}
\usepackage{algorithm}
\usepackage{algorithmic}
\usepackage{pifont}
\usepackage{float}
\usepackage{colortbl}

\usepackage{tikz}
\usetikzlibrary{arrows.meta, positioning, calc, fit, backgrounds, shadows, matrix, decorations.pathreplacing}
\usepackage{pgfplots}
\pgfplotsset{compat=1.18}

\newcommand{\ice}{\textsc{ICE}}
\newcommand{\cmark}{\ding{51}}
\newcommand{\xmark}{\ding{55}}

\title{ICE: Intervention-Consistent Explanation Evaluation \\ with Statistical Grounding for LLMs}

\author{Abhinaba Basu$^{1,2}$ \and Pavan Chakraborty$^{2}$ \\
  $^{1}$National Institute of Electronics and Information Technology, New Delhi, India \\
  $^{2}$Indian Institute of Information Technology Allahabad, Prayagraj, India \\
  \texttt{mail@abhinaba.com}, \texttt{pavan@iiita.ac.in}}

\begin{document}
\maketitle

\begin{abstract}
Evaluating whether explanations faithfully reflect a model's reasoning remains an open problem. Existing benchmarks use single interventions without statistical testing, making it impossible to distinguish genuine faithfulness from chance-level performance. We show that faithfulness is not a fixed property but an \emph{operator-dependent quantity} that changes with the intervention method used to measure it. We introduce \ice{} (Intervention-Consistent Explanation), a framework that evaluates explanations against random baselines of equal size under multiple operators. Evaluating 7 LLMs across 4 tasks with deletion and retrieval infill operators, we find that switching operators crosses the positive-evidence threshold in 18\% of configurations (5 of 28 attention comparisons), with gaps reaching 44 percentage points. Randomized baselines detect anti-faithfulness (explanations worse than random) in nearly one-third of English deletion configurations, invisible without random comparisons. These patterns persist across 6 non-English languages and 2 attribution methods. The methodology generalizes to step-level chain-of-thought evaluation, where preliminary results on 3 frontier models suggest that high accuracy does not imply faithful reasoning.
\end{abstract}

\section{Introduction}

Consider evaluating whether a sentiment model's explanation is faithful. Attention highlights ``gorgeous'' and ``seductive''---the actual sentiment signal. Gradient highlights ``a'' and ``movie''---function words. Which explanation is faithful? The answer seems obvious, but current benchmarks like ERASER~\cite{deyoung2020eraser} would report raw scores for each without telling you whether either outperforms \emph{random tokens}. And the answer depends on a choice the benchmarks treat as incidental: \emph{which intervention operator do you use to test faithfulness?}

This paper argues that faithfulness is operator-dependent, and that evaluation must account for this. While prior work has noted that individual evaluation choices matter~\cite{hase2021ood, kamp2023dynamic}, no framework has systematically quantified operator dependence or provided a protocol for interpreting it.

We first show that different operators introduce different biases (\S\ref{sec:framework}): deletion creates OOD inputs while retrieval infill may preserve competing signals, inflating or deflating faithfulness estimates by up to 44 percentage points. To address this, we introduce \ice{} (Intervention-Consistent Explanation), which evaluates explanations against random baselines of equal size under each operator, yielding win rates with confidence intervals. When operators agree, the evidence for faithfulness is strong; when they disagree, the gap quantifies methodological uncertainty.

Evaluating 7 LLMs across 4 tasks, 2 attribution methods, and 6 non-English languages (\S\ref{sec:results}), we find operator dependence is pervasive. Randomized baselines reveal that explanations perform \emph{worse} than random in nearly one-third of English deletion configurations (18/56)---a finding invisible without random comparisons. The methodology also extends to step-level chain-of-thought evaluation (Appendix~\ref{app:cot}), where preliminary results suggest broad applicability.

ICE is not simply ``re-running ERASER with a second operator.'' It adds randomization testing against size-matched baselines (which no prior framework provides), an integrated protocol for interpreting operator agreement, and retrieval infill as an in-distribution alternative to deletion.

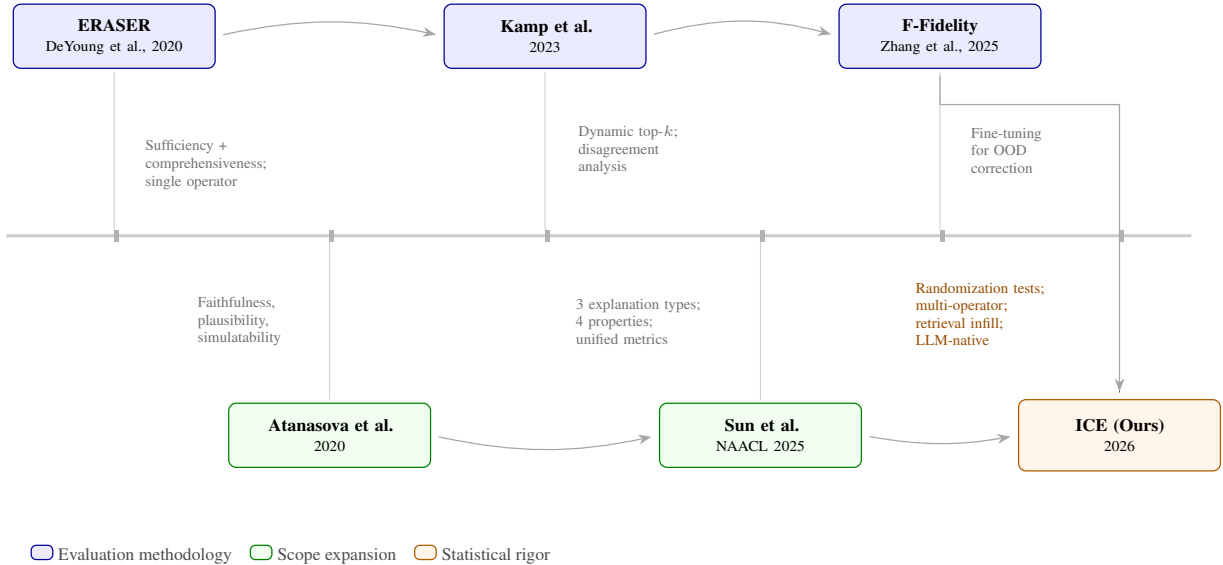
\begin{figure*}[t!]
\centering
\resizebox{\textwidth}{!}{%
\begin{tikzpicture}[
    >=Stealth,
    evalcolor/.style={draw=blue!60!black, fill=blue!8},
    scopecolor/.style={draw=green!50!black, fill=green!6},
    statcolor/.style={draw=orange!70!black, fill=orange!8},
    timenode/.style={
        rounded corners=3pt, line width=0.5pt,
        minimum height=0.9cm, text width=2.6cm, align=center,
        font=\scriptsize, inner sep=3pt,
    },
    annot/.style={
        font=\tiny, text=black!55, align=left, text width=2.4cm
    },
    connector/.style={
        ->, line width=0.5pt, color=black!35, shorten >=3pt, shorten <=3pt
    },
    spine/.style={
        line width=0.4pt, color=black!20
    },
]

\draw[line width=1.2pt, color=black!20] (0, 0) -- (15.5, 0);

\foreach \x in {1.0, 4.0, 7.0, 10.0, 12.5, 15.0} {
    \fill[black!30] (\x, -0.08) rectangle (\x+0.06, 0.08);
}


\node[timenode, evalcolor] (eraser) at (1.0, 2.8) {
    \textbf{ERASER}\\{\tiny DeYoung et al., 2020}
};
\draw[spine] (1.0, 0.08) -- (1.0, 2.3);
\node[annot, anchor=west] at (1.3, 1.0) {Sufficiency +\\comprehensiveness;\\single operator};

\node[timenode, evalcolor] (kamp) at (7.0, 2.8) {
    \textbf{Kamp et al.}\\{\tiny 2023}
};
\draw[spine] (7.0, 0.08) -- (7.0, 2.3);
\node[annot, anchor=east] at (10.0, 1.2) {Dynamic top-$k$;\\disagreement\\analysis};

\node[timenode, evalcolor] (ffid) at (12.5, 2.8) {
    \textbf{F-Fidelity}\\{\tiny Zheng et al., 2025}
};
\draw[spine] (12.5, 0.08) -- (12.5, 2.3);
\node[annot, anchor=west] at (12.8, 1.2) {Fine-tuning\\for OOD\\correction};


\node[timenode, scopecolor] (atan) at (4.0, -2.8) {
    \textbf{Atanasova et al.}\\{\tiny 2020}
};
\draw[spine] (4.0, -0.08) -- (4.0, -2.3);
\node[annot, anchor=east] at (4.7, -1.2) {5 diagnostic\\properties;\\cross-method};

\node[timenode, scopecolor] (sun) at (10.0, -2.8) {
    \textbf{Sun et al.}\\{\tiny NAACL 2025}
};
\draw[spine] (10.0, -0.08) -- (10.0, -2.3);
\node[annot, anchor=west] at (7.3, -1.2) {3 explanation types;\\4 properties;\\unified metrics};

\node[timenode, statcolor, minimum height=1.0cm] (ice) at (15.0, -2.8) {
    \textbf{ICE (Ours)}\\{\tiny 2026}
};
\draw[orange!50!black, line width=0.4pt] (15.0, -0.08) -- (15.0, -2.2);
\node[annot, anchor=east, text=orange!60!black] at (14.7, -1.1) {Randomization tests;\\multi-operator;\\retrieval infill;\\LLM-native};

\draw[connector, bend left=12] (eraser.east) to (kamp.west);
\draw[connector, bend right=12] (atan.east) to (sun.west);
\draw[connector, bend left=15] (kamp.east) to (ffid.west);
\draw[connector, bend right=10] (sun.east) to (ice.west);
\draw[connector] (ffid.south) -- ++(0,-0.5) -| (ice.north);

\node[font=\scriptsize, anchor=north west, text width=14cm] at (0, -4.2) {
    \tikz{\draw[blue!60!black, fill=blue!8, rounded corners=2pt, line width=0.5pt] (0,0) rectangle (0.3,0.2);}\;\textcolor{black!70}{Evaluation}\quad
    \tikz{\draw[green!50!black, fill=green!6, rounded corners=2pt, line width=0.5pt] (0,0) rectangle (0.3,0.2);}\;\textcolor{black!70}{Scope}\quad
    \tikz{\draw[orange!70!black, fill=orange!8, rounded corners=2pt, line width=0.5pt] (0,0) rectangle (0.3,0.2);}\;\textcolor{black!70}{Statistical rigor}
};

\end{tikzpicture}%
}
\caption{
    Evolution of faithfulness evaluation frameworks.
    ICE builds on evaluation methodology
    ({\color{blue!60!black}blue}) and scope expansion ({\color{green!50!black}green}),
    adding statistical rigor ({\color{orange!70!black}orange}) via
    randomization testing and operator-consistent evaluation.
    Arrows show methodological lineage.
}
\label{fig:evolution}
\end{figure*}

\section{Related Work}

\paragraph{Why operator choice matters.}
ERASER~\cite{deyoung2020eraser} established sufficiency and comprehensiveness as standard faithfulness metrics but uses a single deletion operator without statistical testing. Subsequent work has shown that this design conflates faithfulness with OOD degradation~\cite{hase2021ood}, that conclusions depend on top-$k$ selection~\cite{kamp2023dynamic}, and that no single metric works across all tasks~\cite{zaman2025causal}. These findings motivate ICE's core design: evaluate under \emph{multiple} operators with random baselines, and interpret the gap between operators as information about methodological uncertainty.

\paragraph{Prior frameworks.}
\citet{atanasova2020diagnostic} proposed a diagnostic framework evaluating explanations along faithfulness, rationale consistency, and agreement with human annotations. \citet{sun2025unified} compare explanation types under fixed interventions---treating operator choice as incidental. F-Fidelity~\cite{zheng2025ffidelity} addresses OOD via fine-tuning; ICE controls for OOD at the protocol level. \citet{madsen2024faithful} confirm faithfulness depends on explanation type, model, and task, reinforcing multi-method evaluation.

\paragraph{Faithfulness vs.\ plausibility.}
Faithfulness is distinct from plausibility~\cite{jacovi2020faithfully}. We find no detectable correlation between ICE win rates and human rationale alignment ($|r|{<}0.04$), consistent with them measuring different constructs.

\paragraph{Extension to CoT.}
ICE's operator-aware methodology generalizes to step-level CoT evaluation~\cite{turpin2024language, lanham2023measuring}; preliminary results appear in Appendix~\ref{app:cot}.

\section{The \ice{} Framework}
\label{sec:framework}

\paragraph{Terminology.} \textit{Random baselines}: $M{=}50$ random token sets forming the null distribution.
\textit{Empty-input score} $s(\emptyset)$: normalization constant measuring prior label bias.
\textit{Operators}: intervention methods (deletion, retrieval infill).
\textit{Taxonomy}: anti-faithful WR${<}$50\%, near-random 50--55\%, weakly faithful 55--60\%, faithful ${>}$60\%. BH-corrected $p$-values at $\alpha{=}0.10$.

\subsection{Problem Formulation}

Given a model $f$, input $x$ with tokens $(t_1, ..., t_n)$, and an explanation method $E$ producing importance scores $E(x) = (e_1, ..., e_n)$, we evaluate whether the top-$k$ fraction of tokens (rationale $r$, e.g., $k{=}0.2$ retains the top 20\%) identified by $E$ are genuinely important for $f$'s prediction. The central question is whether the explanation method identifies tokens that are more important than randomly selected tokens.

\subsection{NSR and Randomization Testing}

We define Normalized Score Retention (NSR) to measure how much of the original prediction is preserved when rationale tokens are retained and non-rationale tokens are intervened on:

\begin{equation}
\text{NSR}(r) = \frac{s(x_o^r) - s(\emptyset)}{s(x) - s(\emptyset)}
\end{equation}

where $s(x)$ is the prediction score on the original input, $s(x_o^r)$ is the score after retaining rationale $r$ and applying operator $o$ to the non-rationale tokens, and $s(\emptyset)$ is the empty-input baseline. For instruction-tuned models, $s(\emptyset)$ uses the instruction template with empty content (e.g., ``Classify sentiment. Text: [empty] Sentiment:''), measuring prior label bias given task framing alone. For encoders, $s(\emptyset)$ uses [CLS] with an empty sequence. NSR${=}1$ indicates retention equal to the original prediction signal; NSR${=}0$ indicates the empty-input baseline. Values below 0 or above 1 are possible when the intervened score falls below the baseline or exceeds the original.

\paragraph{Concrete Example.}
Consider ``a gorgeous, witty, seductive movie'' classified as positive with $s(x) = 0.78$. Attention highlights \{gorgeous, seductive\} as the rationale $r$. Using deletion as the operator, keeping only these tokens yields $s(x_o^r) = 0.72$. With empty input $s(\emptyset) = 0.50$. Then $\text{NSR} = (0.72 - 0.50)/(0.78 - 0.50) = 0.79$---the rationale preserves 79\% of the prediction signal. Repeating with 50 random token sets: if the rationale beats 46 of 50, the win rate is 92\%, indicating strong faithfulness. Table~\ref{tab:pipeline} shows how operator choice changes this verdict.

\paragraph{The Randomization Test.}

\begin{algorithm}[H]
\caption{ICE Randomization Test}
\footnotesize
\begin{algorithmic}[1]
\REQUIRE Input $x$, rationale $r$, operators $\mathcal{O}$, permutations $M$
\STATE Compute $\text{NSR}_{obs} = \frac{1}{|\mathcal{O}|}\sum_{o \in \mathcal{O}} \text{NSR}_o(r)$
\FOR{$i = 1$ to $M$}
    \STATE Sample random tokens $r_i$ with $|r_i| = |r|$
    \STATE Compute $\text{NSR}_i = \frac{1}{|\mathcal{O}|}\sum_{o \in \mathcal{O}} \text{NSR}_o(r_i)$
\ENDFOR
\STATE \textbf{Win Rate} $= \frac{1}{M}\sum_{i=1}^M \mathbb{1}[\text{NSR}_{obs} > \text{NSR}_i]$
\STATE \textbf{Effect Size} $d_{\text{null}} = \frac{\text{NSR}_{obs} - \mu_{\text{random}}}{\sigma_{\text{random}}}$
\RETURN Win Rate, Effect Size, $p$-value
\end{algorithmic}
\end{algorithm}

\noindent The $p$-value uses a one-sided test:
\[p=\frac{1+\sum_{i=1}^{M}\mathbb{1}[\text{NSR}_i \ge \text{NSR}_{obs}]}{M+1}\]
with $+1$ terms for conservative finite-sample correction. We primarily report \textbf{win rate}, which remains stable even when NSR denominators are small. Unless noted, we report per-operator results rather than the multi-operator average, to reveal operator-specific effects.

\begin{table}[t]
\centering
\footnotesize
\setlength{\tabcolsep}{2pt}
\begin{tabular}{@{}lcc@{}}
\toprule
& \textbf{Delete} & \textbf{Retr.\ Infill} \\
\midrule
Input & \multicolumn{2}{c}{``a gorgeous witty seductive movie''} \\
Rationale & \multicolumn{2}{c}{\{gorgeous, seductive\}} \\
\addlinespace[2pt]
Result & ``gorgeous seductive'' & ``gorgeous \textit{a} seductive \textit{an}'' \\
$s(x_o^r)$ & 0.72 & 0.61 \\
NSR & 0.79 & 0.39 \\
WR & 92\% & 68\% \\
\textbf{Verdict} & \textbf{Faithful} & \textbf{Weakly F.} \\
\bottomrule
\end{tabular}
\caption{ICE pipeline on a sentiment example. Non-rationale tokens are removed (delete) or replaced (infill). The same rationale gets different verdicts.}
\label{tab:pipeline}
\end{table}

\subsection{Operators}

Any intervention creates inputs the model was not trained on. Different interventions introduce different OOD artifacts. Using a single operator risks confounding faithfulness with operator-specific degradation \cite{hase2021ood}. \ice{} employs multiple operators:

\begin{itemize}
    \item \textbf{Deletion}: Removes tokens from the sequence. Fast but produces unnatural truncated text.
    \item \textbf{Retrieval Infill}: Replaces non-rationale tokens with tokens sampled from other examples, excluding the current example (leave-one-out) and filtering label-indicative tokens (label-blacklisted). Preserves input length while destroying task-relevant content. Replacement tokens are drawn from the same corpus, which may introduce label-correlated artifacts; label blacklisting mitigates but does not eliminate this risk.
\end{itemize}

\noindent For encoder models, we additionally evaluate mask-based operators (Appendix~\ref{app:operator_ablation}). For autoregressive LLMs, masking creates degenerate outputs; we use deletion and retrieval infill.

\subsection{Statistical Framework}

We compute 95\% bootstrap confidence intervals ($B=200$ resamples) for uncertainty quantification and apply Benjamini-Hochberg FDR correction ($\alpha = 0.10$) for multiple testing. Full details appear in Appendix~\ref{app:stats}.

\section{Experimental Setup}

We focus on autoregressive LLMs; encoder and Chain-of-Thought evaluations appear in Appendices~\ref{app:encoder} and~\ref{app:cot}.

We evaluate 7 LLMs (1.5B--8B parameters; GPT-2, Llama~3.x, Qwen~2.5, Mistral, DeepSeek, LFM2) on 4 English datasets (SST-2, IMDB, e-SNLI, AG~News) and 6 non-English languages using native sentiment data (French, German, Hindi, Chinese, Turkish, Arabic---covering 4 scripts and 4 language families).
We compare attention and gradient attribution methods.
Full model and dataset details appear in Appendix~\ref{app:models_datasets}.

We use $k=0.2$ (top 20\% tokens), $N=500$ examples per dataset, $M=50$ permutations for LLMs ($M=100$ for encoders), and 512-token truncation. A $k$-sweep in Appendix~\ref{app:ksweep} justifies $k=0.2$ as a middle ground. We release ICEBench with pinned dataset versions for reproducibility (Appendix~\ref{app:models_datasets}).

\section{Results}
\label{sec:results}

We present results that develop the central argument---faithfulness is operator-dependent---in three stages: English benchmarks establish the phenomenon (\S\ref{sec:english}), operator comparison quantifies it (\S\ref{sec:retrieval_comparison}), and multilingual evaluation confirms it generalizes (\S\ref{sec:multilingual}).

\subsection{English Benchmark Results}
\label{sec:english}


\begin{figure*}[t]
\centering
\resizebox{\textwidth}{!}{%
\begin{tikzpicture}[
    cell/.style={
        minimum width=1.15cm, minimum height=0.62cm,
        align=center, font=\scriptsize,
        inner sep=0pt, outer sep=0pt,
    },
    header/.style={
        minimum width=1.15cm, minimum height=0.5cm,
        align=center, font=\scriptsize\bfseries,
        inner sep=1pt, outer sep=0pt, text=black!70,
    },
    rowlabel/.style={
        minimum width=2.0cm, minimum height=0.62cm,
        align=right, font=\scriptsize,
        inner sep=3pt, outer sep=0pt, text=black!70,
        anchor=east,
    },
    paneltitle/.style={
        font=\footnotesize\bfseries, text=black!70,
    },
]

\def\cw{1.15}  
\def\ch{0.62}  
\def\hch{0.31} 


\newcommand{\hcell}[4]{%
    \fill[#3] (#1-\cw/2, #2-\hch) rectangle (#1+\cw/2, #2+\hch);
    \draw[black!20, line width=0.2pt] (#1-\cw/2, #2-\hch) rectangle (#1+\cw/2, #2+\hch);
    \node[cell] at (#1, #2) {#4};
}

\def\rg{1.66}   
\def\ra{1.06}   
\def\rb{0.46}   
\def\rc{-0.14}  
\def\rd{-0.74}  
\def\re{-1.34}  
\def\rf{-1.94}  

\node[paneltitle] at (1.72, 2.42) {Delete + Attention};

\node[header] at (0.0, 2.08) {SST-2};
\node[header] at (\cw, 2.08) {IMDB};
\node[header] at (2*\cw, 2.08) {e-SNLI};
\node[header] at (3*\cw, 2.08) {AG};

\node[rowlabel] at (-0.2, \rg) {GPT-2};
\node[rowlabel] at (-0.2, \ra) {LFM2};
\node[rowlabel] at (-0.2, \rb) {Llama-3.2};
\node[rowlabel] at (-0.2, \rc) {Llama-3.1};
\node[rowlabel] at (-0.2, \rd) {Qwen};
\node[rowlabel] at (-0.2, \re) {Mistral};
\node[rowlabel] at (-0.2, \rf) {DeepSeek};

\hcell{0.0}{\rg}{green!35!white}{\textbf{60.6}}
\hcell{\cw}{\rg}{yellow!25!white}{44.0}
\hcell{2*\cw}{\rg}{green!35!white}{\textbf{64.0}}
\hcell{3*\cw}{\rg}{green!35!white}{\textbf{70.8}}

\hcell{0.0}{\ra}{yellow!25!white}{45.4}
\hcell{\cw}{\ra}{yellow!25!white}{50.3}
\hcell{2*\cw}{\ra}{green!35!white}{\textbf{66.6}}
\hcell{3*\cw}{\ra}{green!20!white}{57.0}

\hcell{0.0}{\rb}{green!20!white}{53.2}
\hcell{\cw}{\rb}{green!35!white}{\textbf{71.3}}
\hcell{2*\cw}{\rb}{green!35!white}{\textbf{86.4}}
\hcell{3*\cw}{\rb}{green!20!white}{56.0}

\hcell{0.0}{\rc}{green!20!white}{53.2}
\hcell{\cw}{\rc}{green!20!white}{52.5}
\hcell{2*\cw}{\rc}{green!35!white}{\textbf{85.2}}
\hcell{3*\cw}{\rc}{yellow!25!white}{47.5}

\hcell{0.0}{\rd}{green!35!white}{\textbf{68.4}}
\hcell{\cw}{\rd}{green!35!white}{\textbf{94.9}}
\hcell{2*\cw}{\rd}{green!35!white}{\textbf{77.3}}
\hcell{3*\cw}{\rd}{green!35!white}{\textbf{62.2}}

\hcell{0.0}{\re}{green!20!white}{57.5}
\hcell{\cw}{\re}{green!35!white}{\textbf{83.8}}
\hcell{2*\cw}{\re}{green!35!white}{\textbf{71.1}}
\hcell{3*\cw}{\re}{green!20!white}{51.7}

\hcell{0.0}{\rf}{green!35!white}{\textbf{62.2}}
\hcell{\cw}{\rf}{green!35!white}{\textbf{84.6}}
\hcell{2*\cw}{\rf}{green!35!white}{\textbf{63.7}}
\hcell{3*\cw}{\rf}{yellow!25!white}{49.4}

\begin{scope}[xshift=6.2cm]

\node[paneltitle] at (1.72, 2.42) {Retrieval + Attention};

\node[header] at (0.0, 2.08) {SST-2};
\node[header] at (\cw, 2.08) {IMDB};
\node[header] at (2*\cw, 2.08) {e-SNLI};
\node[header] at (3*\cw, 2.08) {AG};

\hcell{0.0}{\rg}{green!20!white}{52.4}
\hcell{\cw}{\rg}{red!30!white}{38.3}
\hcell{2*\cw}{\rg}{green!35!white}{\textbf{68.2}}
\hcell{3*\cw}{\rg}{green!35!white}{\textbf{73.2}}

\hcell{0.0}{\ra}{yellow!25!white}{45.5}
\hcell{\cw}{\ra}{yellow!25!white}{49.4}
\hcell{2*\cw}{\ra}{green!35!white}{\textbf{65.5}}
\hcell{3*\cw}{\ra}{green!20!white}{51.1}

\hcell{0.0}{\rb}{green!20!white}{52.7}
\hcell{\cw}{\rb}{green!35!white}{\textbf{91.8}}
\hcell{2*\cw}{\rb}{red!30!white}{42.6}
\hcell{3*\cw}{\rb}{green!20!white}{56.9}

\hcell{0.0}{\rc}{green!20!white}{51.5}
\hcell{\cw}{\rc}{green!35!white}{\textbf{75.3}}
\hcell{2*\cw}{\rc}{green!35!white}{\textbf{74.0}}
\hcell{3*\cw}{\rc}{yellow!25!white}{49.4}

\hcell{0.0}{\rd}{green!20!white}{59.6}
\hcell{\cw}{\rd}{green!35!white}{\textbf{96.3}}
\hcell{2*\cw}{\rd}{green!35!white}{\textbf{77.5}}
\hcell{3*\cw}{\rd}{green!20!white}{59.8}

\hcell{0.0}{\re}{green!20!white}{56.1}
\hcell{\cw}{\re}{green!35!white}{\textbf{89.8}}
\hcell{2*\cw}{\re}{green!35!white}{\textbf{67.5}}
\hcell{3*\cw}{\re}{red!30!white}{42.7}

\hcell{0.0}{\rf}{green!20!white}{53.9}
\hcell{\cw}{\rf}{green!35!white}{\textbf{80.5}}
\hcell{2*\cw}{\rf}{green!35!white}{\textbf{65.5}}
\hcell{3*\cw}{\rf}{yellow!25!white}{46.7}

\end{scope}

\begin{scope}[yshift=-5.2cm]

\node[paneltitle] at (1.72, 2.42) {Delete + Gradient};

\node[header] at (0.0, 2.08) {SST-2};
\node[header] at (\cw, 2.08) {IMDB};
\node[header] at (2*\cw, 2.08) {e-SNLI};
\node[header] at (3*\cw, 2.08) {AG};

\node[rowlabel] at (-0.2, \rg) {GPT-2};
\node[rowlabel] at (-0.2, \ra) {LFM2};
\node[rowlabel] at (-0.2, \rb) {Llama-3.2};
\node[rowlabel] at (-0.2, \rc) {Llama-3.1};
\node[rowlabel] at (-0.2, \rd) {Qwen};
\node[rowlabel] at (-0.2, \re) {Mistral};
\node[rowlabel] at (-0.2, \rf) {DeepSeek};

\hcell{0.0}{\rg}{green!20!white}{56.0}
\hcell{\cw}{\rg}{red!30!white}{42.6}
\hcell{2*\cw}{\rg}{red!30!white}{29.5}
\hcell{3*\cw}{\rg}{yellow!25!white}{48.7}

\hcell{0.0}{\ra}{red!30!white}{41.8}
\hcell{\cw}{\ra}{green!20!white}{54.0}
\hcell{2*\cw}{\ra}{yellow!25!white}{45.3}
\hcell{3*\cw}{\ra}{yellow!25!white}{49.6}

\hcell{0.0}{\rb}{red!30!white}{42.4}
\hcell{\cw}{\rb}{green!35!white}{\textbf{68.6}}
\hcell{2*\cw}{\rb}{green!35!white}{\textbf{83.7}}
\hcell{3*\cw}{\rb}{red!30!white}{44.2}

\hcell{0.0}{\rc}{yellow!25!white}{46.0}
\hcell{\cw}{\rc}{yellow!25!white}{47.4}
\hcell{2*\cw}{\rc}{green!35!white}{\textbf{97.2}}
\hcell{3*\cw}{\rc}{red!30!white}{42.2}

\hcell{0.0}{\rd}{green!20!white}{55.4}
\hcell{\cw}{\rd}{green!35!white}{\textbf{91.4}}
\hcell{2*\cw}{\rd}{green!20!white}{53.9}
\hcell{3*\cw}{\rd}{green!20!white}{55.1}

\hcell{0.0}{\re}{yellow!25!white}{47.4}
\hcell{\cw}{\re}{green!35!white}{\textbf{78.4}}
\hcell{2*\cw}{\re}{yellow!25!white}{50.7}
\hcell{3*\cw}{\re}{green!35!white}{\textbf{60.7}}

\hcell{0.0}{\rf}{red!30!white}{40.5}
\hcell{\cw}{\rf}{green!35!white}{\textbf{70.2}}
\hcell{2*\cw}{\rf}{green!20!white}{55.2}
\hcell{3*\cw}{\rf}{red!30!white}{39.1}

\end{scope}

\begin{scope}[xshift=6.2cm, yshift=-5.2cm]

\node[paneltitle] at (1.72, 2.42) {Retrieval + Gradient};

\node[header] at (0.0, 2.08) {SST-2};
\node[header] at (\cw, 2.08) {IMDB};
\node[header] at (2*\cw, 2.08) {e-SNLI};
\node[header] at (3*\cw, 2.08) {AG};

\hcell{0.0}{\rg}{red!30!white}{44.1}
\hcell{\cw}{\rg}{red!30!white}{37.2}
\hcell{2*\cw}{\rg}{red!30!white}{24.8}
\hcell{3*\cw}{\rg}{red!30!white}{42.6}

\hcell{0.0}{\ra}{red!30!white}{43.7}
\hcell{\cw}{\ra}{green!20!white}{51.3}
\hcell{2*\cw}{\ra}{green!20!white}{52.9}
\hcell{3*\cw}{\ra}{red!30!white}{43.9}

\hcell{0.0}{\rb}{yellow!25!white}{45.9}
\hcell{\cw}{\rb}{green!35!white}{\textbf{92.6}}
\hcell{2*\cw}{\rb}{green!35!white}{\textbf{83.4}}
\hcell{3*\cw}{\rb}{red!30!white}{36.9}

\hcell{0.0}{\rc}{yellow!25!white}{47.7}
\hcell{\cw}{\rc}{green!35!white}{\textbf{78.5}}
\hcell{2*\cw}{\rc}{green!35!white}{\textbf{90.1}}
\hcell{3*\cw}{\rc}{red!30!white}{44.4}

\hcell{0.0}{\rd}{green!20!white}{53.8}
\hcell{\cw}{\rd}{green!35!white}{\textbf{95.0}}
\hcell{2*\cw}{\rd}{green!20!white}{57.7}
\hcell{3*\cw}{\rd}{yellow!25!white}{48.6}

\hcell{0.0}{\re}{yellow!25!white}{49.5}
\hcell{\cw}{\re}{green!35!white}{\textbf{87.0}}
\hcell{2*\cw}{\re}{green!20!white}{51.3}
\hcell{3*\cw}{\re}{green!20!white}{53.8}

\hcell{0.0}{\rf}{yellow!25!white}{48.0}
\hcell{\cw}{\rf}{green!35!white}{\textbf{87.7}}
\hcell{2*\cw}{\rf}{green!35!white}{\textbf{63.1}}
\hcell{3*\cw}{\rf}{red!30!white}{33.7}

\end{scope}

\draw[black!15, dashed, line width=0.4pt] (5.2, 2.7) -- (5.2, -7.3);  
\draw[black!15, dashed, line width=0.4pt] (-2.1, -2.55) -- (11.0, -2.55);  

\begin{scope}[yshift=-8.6cm, xshift=2.0cm]
    \shade[left color=red!30!white, right color=yellow!25!white]
        (0, 0) rectangle (2.5, 0.22);
    \shade[left color=yellow!25!white, right color=green!35!white]
        (2.5, 0) rectangle (5.0, 0.22);
    \draw[black!25, line width=0.2pt] (0, 0) rectangle (5.0, 0.22);
    \node[font=\tiny, text=black!55, below] at (0, -0.02) {$<$40\%};
    \node[font=\tiny, text=black!55, below] at (1.25, -0.02) {Anti-faithful};
    \node[font=\tiny, text=black!55, below] at (2.5, -0.02) {50\%};
    \node[font=\tiny, text=black!55, below] at (3.75, -0.02) {Faithful};
    \node[font=\tiny, text=black!55, below] at (5.0, -0.02) {$>$60\%};
\end{scope}

\draw[black!15, line width=0.3pt, rounded corners=2pt]
    (-2.1, -9.2) rectangle (11.0, 2.8);

\end{tikzpicture}%
}
\caption{
    English faithfulness (win rate \%) under both operators and both attribution methods.
    Top: Attention. Bottom: Gradient. Left: Deletion. Right: Retrieval Infill.
    Models (top to bottom): GPT-2, LFM2, Llama-3.2, Llama-3.1, Qwen, Mistral, DeepSeek.
    Right panels share the same model order.
    On short text (SST-2), deletion yields higher estimates; on IMDB (long text), the pattern reverses for most models.
    Green = faithful ($>$60\%), yellow = near-random, red = strongly anti-faithful ($<$40\%); values below 50\% are anti-faithful per the taxonomy in \S3.
}
\label{fig:english_heatmap}
\end{figure*}
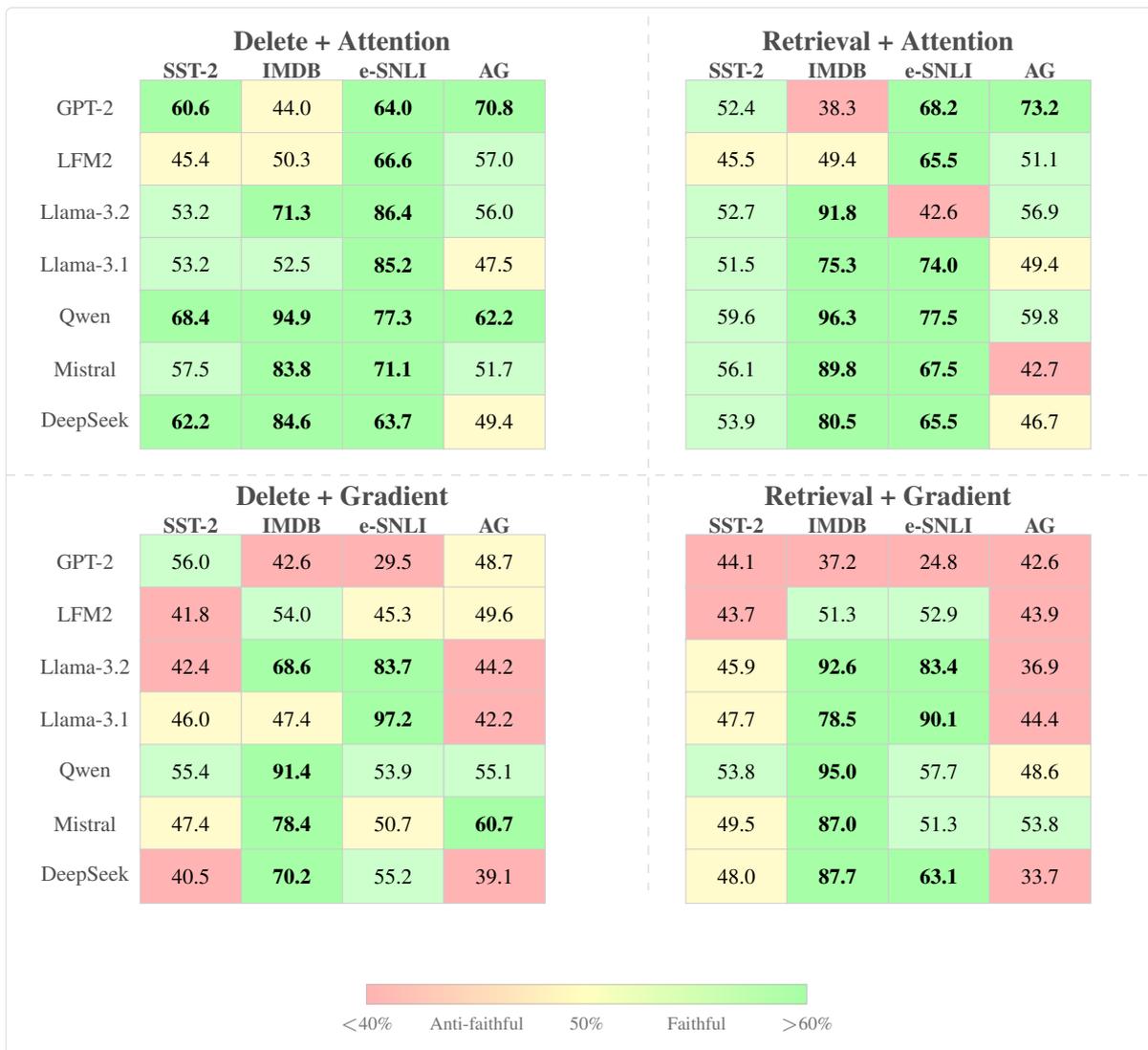

Figure~\ref{fig:english_heatmap} shows win rates under both operators (deletion and retrieval infill) and both attribution methods, revealing four patterns.

\paragraph{Operator effects.} Comparing left (deletion) and right (retrieval infill) panels reveals systematic differences. On short text, deletion produces more green cells, particularly on SST-2 (gaps of 8--9\,pp) and e-SNLI (up to 44\,pp for Llama-3.2). On IMDB (long text), the pattern reverses for most models---retrieval infill yields \emph{higher} win rates than deletion (e.g., Llama-3.2: 91.8\% vs.\ 71.3\%), likely because deleting most of a long review creates severely degraded input, while retrieval preserves natural text length. This suggests the operator gap direction is text-length sensitive.

\paragraph{Short vs.\ Long Text.} Under both operators, attention beats gradient on short text (SST-2). Both converge on long text (IMDB: ${\sim}$85--95\% for capable models under both operators). Attention captures local sentiment signals; gradient benefits from accumulated context.

\paragraph{Task-Specific Patterns.} NLI favors attention under deletion (Llama 3.2: 86.4\%) but shows more nuanced patterns under retrieval infill (Llama 3.2 attention drops to 42.6\%, while gradient rises to 83.4\%). Topic classification generally favors attention, with Mistral as the main exception.

\paragraph{Base vs.\ Instruct.} Llama 3.1-8B Base shows task-dependent faithfulness under both operators: near-random on SST-2 but exceptional on e-SNLI (97.2\% gradient under deletion, 90.1\% under retrieval---the highest across all configurations).

\subsection{Multilingual Results}
\label{sec:multilingual}

\begin{figure*}[t]
\centering
\includegraphics[width=0.95\textwidth]{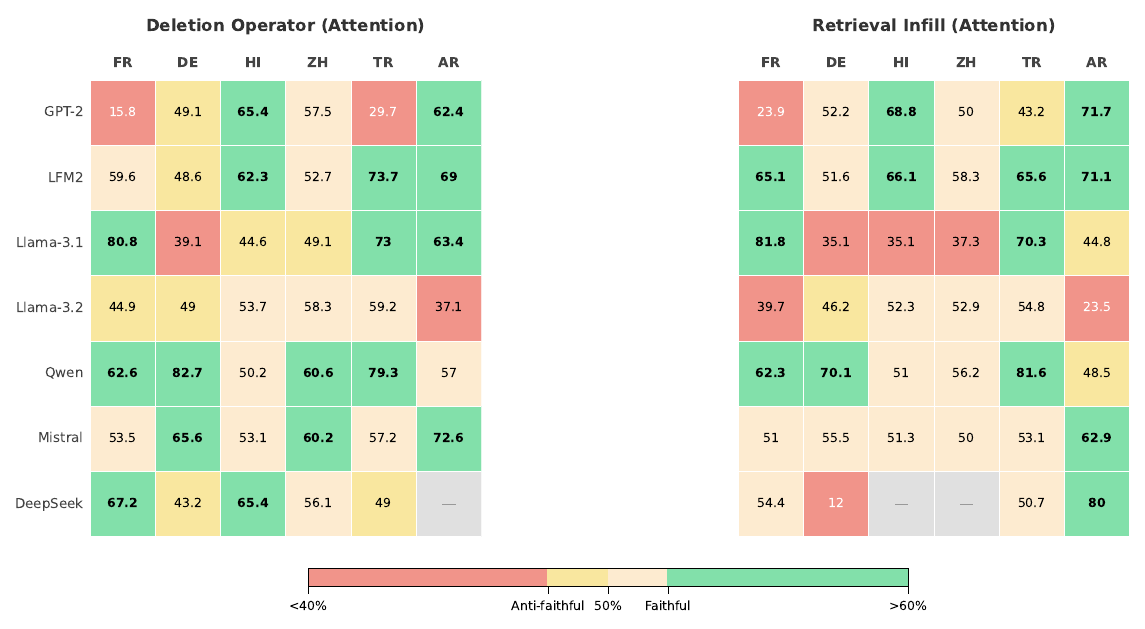}
\caption{Multilingual faithfulness (attention win rate \%) under both operators across 6 languages and 4 scripts. \textbf{Left:} Deletion. \textbf{Right:} Retrieval Infill. The color scale highlights strongly anti-faithful values ($<$40\%); values below 50\% are anti-faithful under the taxonomy in \S3. Gray = no valid output.}
\label{fig:multilingual_heatmap}
\end{figure*}

Figure~\ref{fig:multilingual_heatmap} reveals cross-lingual variation spanning 12.0\% to 82.7\% across both operators. No single model dominates under deletion: Qwen 2.5-7B leads on German (82.7\%) and Turkish (79.3\%), Llama 3.1 on French (80.8\%), while GPT-2 shows anti-faithfulness on French (15.8\%) and Turkish (29.7\%). Tokenization does not predict faithfulness: GPT-2 achieves 66\% Hindi gradient despite 8.1$\times$ token expansion, while French (1.8$\times$) yields anti-faithful results (15.8\%).

Under retrieval infill (right panel), the operator gap varies by language: GPT-2 Hindi \emph{increases} from 65.4\% to 68.8\%---retrieval exceeding deletion, as also observed on IMDB for most models---while Llama 3.1 drops to anti-faithful (35.1\% Hindi, 37.3\% Chinese). Overall, cross-lingual faithfulness is not explained by tokenization alone. The mechanism is that over-segmented tokenizers split semantically coherent units across multiple tokens, diluting per-token attribution scores and creating misalignment between token boundaries and linguistic units~\cite{rust2021good}. However, model-language interactions, morphological structure, and operator sensitivity jointly determine whether attributions remain faithful---tokenization is one factor among several.

\paragraph{Translation-paired control.}
A concern with comparing faithfulness across languages is that native datasets differ in content, confounding language with dataset effects. We control for this using NLLB-200~\cite{nllb2022} machine translation to translate English SST-2 into French and German ($N{=}200$):

\begin{table}[h]
\centering
\small
\begin{tabular}{lcc}
\toprule
\textbf{Condition} & \textbf{French} & \textbf{German} \\
\midrule
GPT-2 (native) & 15.8\% & 52.4\% \\
GPT-2 (translated) & 37.6\% & 34.1\% \\
Qwen2.5-3B (translated) & 36.3\% & 41.4\% \\
\bottomrule
\end{tabular}
\caption{Translation-paired control: same source text, different languages. Native vs.\ translated results differ substantially (GPT-2 German: 52.4\% native vs.\ 34.1\% translated), confirming that absolute cross-language rankings reflect dataset--language interactions, not pure language effects.}
\label{tab:nllb}
\end{table}

\noindent This control narrows our multilingual claim: we report faithfulness as dependent on model--language--dataset--operator interactions, not as a language ranking. Full language-specific analysis appears in Appendix~\ref{app:language_patterns}.

\subsection{Effect Sizes and Anti-Faithfulness}

Effect sizes quantify faithfulness magnitude beyond win rates. Llama 3.1-8B e-SNLI gradient achieves $d_{\text{null}} = 2.50$ (2.5 standard deviations above the random baseline), while GPT-2 French gradient shows $d_{\text{null}} = -2.36$ (severe anti-faithfulness). We find anti-faithfulness (win rate $<$ 50\%) in nearly one-third of English deletion configurations (18 of 56), predominantly gradient-based, where gradient assigns highest importance to sentence-initial function words while ignoring task-relevant content. Anti-faithful explanations actively mislead users; practitioners must verify faithfulness on their specific configuration. Full effect size tables and bootstrap CIs appear in Appendix~\ref{app:effect_sizes}.

\section{Analysis}

\subsection{Faithfulness vs.\ Plausibility}

\begin{figure*}[t]
\centering
\includegraphics[width=0.85\textwidth]{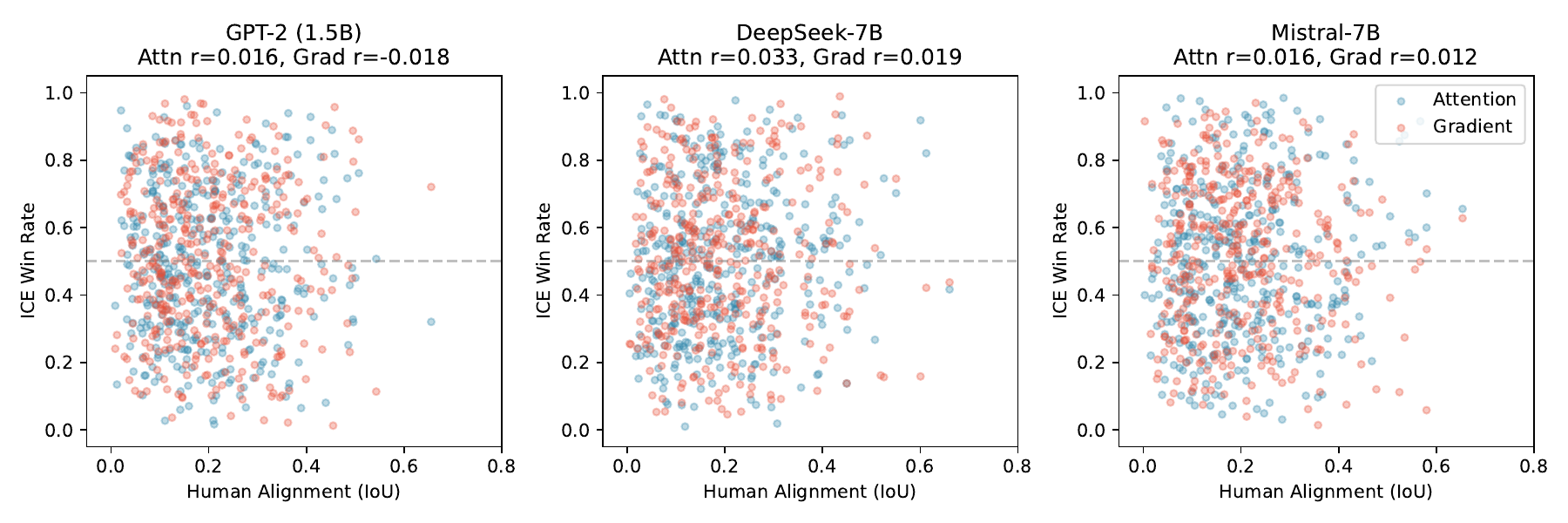}
\caption{IoU (human alignment, x-axis) vs.\ ICE Win Rate (faithfulness, y-axis) across GPT-2, DeepSeek, and Mistral on e-SNLI. All $|r| < 0.04$: no correlation between human alignment and computational faithfulness.}
\label{fig:plausibility}
\end{figure*}

A key question is whether operator-dependent faithfulness correlates with human plausibility judgments. We measure plausibility as IoU between attributed tokens and human-annotated rationales: $\text{IoU} = |R_{\text{attr}} \cap R_{\text{human}}| / |R_{\text{attr}} \cup R_{\text{human}}|$ at $k{=}0.2$. All models show near-zero Pearson correlation between IoU and ICE win rate ($|r| < 0.04$, $p > 0.5$; Figure~\ref{fig:plausibility}). With $N \approx 400$ per model, we have approximately 80\% power to detect $r \geq 0.14$, so we cannot rule out small correlations. However, the consistency across 1.5B--8B models on e-SNLI suggests no detectable relationship at this sample size. Plausibility benchmarks do not appear to measure faithfulness; both likely require independent assessment (full statistics in Appendix~\ref{app:alignment}).

\subsection{Retrieval Infill: Does Operator Choice Matter?}
\label{sec:retrieval_comparison}

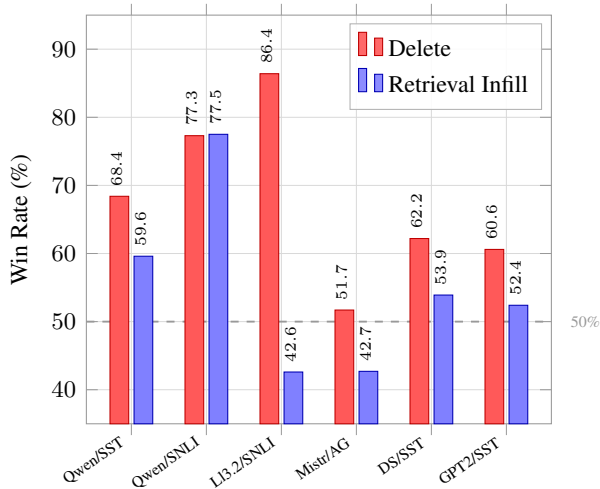
\begin{figure}[t]
\centering
\begin{tikzpicture}
\begin{axis}[
    width=\columnwidth,
    height=7.0cm,
    ybar,
    bar width=7pt,
    ymin=35, ymax=95,
    ylabel={Win Rate (\%)},
    ylabel style={font=\small},
    symbolic x coords={
        {Qwen/SST},
        {Qwen/SNLI},
        {Ll3.2/SNLI},
        {Mistr/AG},
        {DS/SST},
        {GPT2/SST}
    },
    xtick=data,
    xticklabel style={font=\tiny, rotate=35, anchor=east},
    yticklabel style={font=\footnotesize},
    ytick={40,50,60,70,80,90},
    grid=major,
    grid style={line width=0.2pt, draw=black!12},
    major grid style={line width=0.3pt, draw=black!15},
    axis line style={draw=black!40},
    tick style={draw=black!40},
    enlarge x limits=0.12,
    legend style={
        at={(0.98,0.98)},
        anchor=north east,
        font=\footnotesize,
        draw=black!30,
        fill=white,
        fill opacity=0.9,
        text opacity=1,
        row sep=1pt,
        inner sep=3pt,
        legend cell align=left,
    },
    nodes near coords,
    every node near coord/.append style={font=\tiny, rotate=90, anchor=west, xshift=0pt, yshift=1pt},
    clip=false,
]

\draw[dashed, black!40, line width=0.7pt] (rel axis cs:0,{(50-35)/(95-35)}) -- (rel axis cs:1,{(50-35)/(95-35)});

\addplot[
    fill=red!65!white,
    draw=red!75!black,
    line width=0.4pt,
] coordinates {
    ({Qwen/SST}, 68.4)
    ({Qwen/SNLI}, 77.3)
    ({Ll3.2/SNLI}, 86.4)
    ({Mistr/AG}, 51.7)
    ({DS/SST}, 62.2)
    ({GPT2/SST}, 60.6)
};

\addplot[
    fill=blue!50!white,
    draw=blue!70!black,
    line width=0.4pt,
] coordinates {
    ({Qwen/SST}, 59.6)
    ({Qwen/SNLI}, 77.5)
    ({Ll3.2/SNLI}, 42.6)
    ({Mistr/AG}, 42.7)
    ({DS/SST}, 53.9)
    ({GPT2/SST}, 52.4)
};

\legend{Delete, Retrieval Infill}

\node[font=\tiny, text=black!40, anchor=west] at (rel axis cs:1.02,{(50-35)/(95-35)}) {50\%};

\end{axis}
\end{tikzpicture}
\caption{
    Operator comparison across model-dataset configurations (attention, main models).
    Delete ({\color{red!75!black}red}) typically yields higher win rates on short text,
    while Retrieval Infill ({\color{blue!70!black}blue}) provides conservative estimates.
    The gap reaches 43.8\,pp (Llama-3.2/e-SNLI), but operators \emph{agree} on Qwen/e-SNLI (77.3\% vs.\ 77.5\%).
    Representative comparisons in Appendix Table~\ref{tab:retrieval_comparison_appendix}; full attention results in Appendix Table~\ref{tab:english_results_appendix}.
}
\label{fig:operator_both}
\end{figure}

As shown in the 4-panel comparison (Figure~\ref{fig:english_heatmap}), operator choice substantially changes win rates, with deletion typically exceeding retrieval on short text but the pattern reversing on long text. Figure~\ref{fig:operator_both} quantifies this effect on representative configurations.


The results confirm that operator choice fundamentally changes conclusions. Delete classifies Llama-3.2 e-SNLI as faithful (86.4\%), but Retrieval Infill downgrades it to anti-faithful (42.6\%)---a 44 percentage point gap. Conversely, both operators \emph{agree} on Qwen e-SNLI (${\sim}$77\%), providing stronger evidence of genuine faithfulness there.

Across all 28 attention configurations (7 models $\times$ 4 datasets), switching operators crosses the 55\% positive-evidence threshold (from weakly-faithful-or-better to near-random-or-worse, or vice versa) in 5 cases (18\%), with the most extreme gap reaching 44\,pp. Deletion exceeds retrieval in 67\% of short-text cases (median gap 1.7\,pp, mean 4.6\,pp). On long text (IMDB), the relationship reverses for most models. We recommend reporting both operators: when operators agree, the evidence is strong regardless of which is higher. When they disagree, the gap quantifies methodological uncertainty and practitioners should treat both estimates as informative bounds. Full attention results appear in Appendix~\ref{app:english_results}; representative paired comparisons in Appendix~\ref{app:retrieval}.

\paragraph{Validating operator agreement on known-outcome cases.}
To validate that operator agreement is a reliable signal, we examine cases with independently verifiable outcomes. For anti-faithful attributions---where gradient selects ``a'' instead of sentiment adjectives like ``gorgeous'' (Table~\ref{tab:anti_faith_examples})---both operators correctly assign WR\,=\,0\% across all 4 examined cases. For genuinely faithful cases (Llama~3.1 e-SNLI gradient, Qwen IMDB attention), both operators converge at WR\,$>$\,77\% with gaps $<$\,8\,pp. This pattern---agreement on clear positives and clear negatives---validates operator convergence as a calibration signal.

\paragraph{Comparison with F-Fidelity.}
F-Fidelity \cite{zheng2025ffidelity} addresses OOD artifacts by fine-tuning the model to adapt to perturbed inputs. ICE instead controls for OOD at the input level via retrieval infill, preserving the model unchanged. The approaches answer different questions: F-Fidelity asks ``would this model be faithful if it understood perturbed inputs?'' while ICE asks ``is this model faithful as deployed?'' For deployment auditing, evaluating the unmodified model is the relevant question. F-Fidelity also requires per-dataset fine-tuning (hours per configuration), while ICE is training-free (minutes), enabling the cross-lingual and multi-model comparisons central to this paper.

\subsection{Head-to-Head Comparison with ERASER}

To quantify ICE's practical advantage over existing frameworks, we evaluate both ERASER and ICE on 120 step-level CoT configurations (10 frontier API models $\times$ 6 reasoning tasks $\times$ 2 metrics; details in supplementary materials). We threshold ERASER's comprehensiveness at 0.5 and ICE's win rate at 55\% for categorical comparison.

\begin{table}[t]
\centering
\caption{ERASER-style comprehensiveness vs ICE. In 49/53 disagreements (92\%), ERASER yields a positive verdict where ICE does not.}
\label{tab:eraser-comparison}
\footnotesize
\setlength{\tabcolsep}{2.5pt}
\begin{tabular}{llccc}
\toprule
\textbf{Model / Data} & \textbf{Met.} & \textbf{Raw} & \textbf{WR} & \textbf{Agree?} \\
\midrule
\multicolumn{5}{l}{\textit{ERASER-positive, ICE-negative (shaded)}} \\
\rowcolor{red!8} Llama-70B / CSQA & C & .95 & .19 & \xmark \\
\rowcolor{red!8} Llama-70B / SST & C & .86 & .20 & \xmark \\
\rowcolor{red!8} Llama-70B / AG & C & .96 & .36 & \xmark \\
\rowcolor{red!8} Qwen-32B / GSM & C & .97 & .42 & \xmark \\
\rowcolor{red!8} Grok4 / ARC & S & .75 & .12 & \xmark \\
\rowcolor{red!8} DSv3 / MedQA & S & .79 & .21 & \xmark \\
\midrule
\multicolumn{5}{l}{\textit{Both frameworks agree}} \\
Gemini3 / GSM & C & .73 & .56 & \cmark \\
Grok4 / MedQA & C & .48 & .51 & \cmark \\
Kimi / MedQA & C & .48 & .53 & \cmark \\
\bottomrule
\end{tabular}
\end{table}

Table~\ref{tab:eraser-comparison} shows representative configurations. Across all 120, the two frameworks disagree on comprehensiveness in 44\% of cases, with ERASER yielding a positive verdict where ICE does not in 92\% of disagreements. The core issue is that ERASER's raw scores are typically high (0.65--0.99) in disagreement cases because removing tokens always degrades output, but ICE reveals that random tokens produce similarly high scores---the apparent ``faithfulness'' is an artifact of input degradation, not explanation quality.

\subsection{Contamination Audit for Retrieval Infill}

A concern with retrieval infill is that replacement tokens drawn from the same corpus may contain label-adjacent tokens that survive the blacklist, inflating sufficiency estimates. We conducted a full-corpus audit across three datasets:

\begin{table}[h]
\centering
\small
\begin{tabular}{lccc}
\toprule
\textbf{Dataset} & \textbf{Blocked} & \textbf{Surviving} & \textbf{Est.\ rate} \\
\midrule
SST-2 & 49 & 25 (subword) & 0.20\% \\
AG News & 0 & 18 (full words) & 0.14\% \\
e-SNLI & 13 & 9 (prefixes) & 0.06\% \\
\bottomrule
\end{tabular}
\caption{Contamination audit: tokens surviving the blacklist per dataset. AG News topic words are not in the current blacklist---a gap we acknowledge. However, the theoretical contamination rate per example is $< 0.2\%$ (probability of sampling a surviving token in $k{=}4$ random draws from 50K vocabulary).}
\label{tab:contamination}
\end{table}

Empirical validation on 490 examples found 3.7\% contamination at a coarser level (checking full infilled token sequences), with $\Delta$WR$=3$pp between contaminated and clean subsets---not statistically significant. We recommend adding topic classification words to the blacklist in future work.

\subsection{Multilingual Gradient Results}

Figure~\ref{fig:multilingual_heatmap} reports only attention for the cross-lingual analysis. We additionally evaluate gradient attribution for all 7 models across 6 languages (complete results in supplementary material; language-specific patterns in Appendix~\ref{app:language_patterns}). Key divergences: LFM2 French gradient 73.4\% vs.\ attention 59.6\% (gradient more faithful); Qwen German attention 82.7\% vs.\ gradient 80.4\% (close agreement). The cross-lingual conclusion---operator dependence persists across languages---holds under both attribution methods.

\subsection{Practical Considerations}

Instruction-tuned models show higher sentiment faithfulness than base models, likely from alignment training. LFM2-2.6B shows task-specific behavior---high NLI but near-random sentiment---suggesting efficient architectures trade broad faithfulness for task-specific capability.

Prompt sensitivity analysis (Appendix~\ref{app:prompt_sensitivity}) reveals that some prompts induce anti-faithfulness and that confidence does not predict faithfulness. Long text reduces method sensitivity ($|\Delta| < 2.1\%$ on IMDB vs.\ 11.8\% on SST-2). Our practical guidelines for selecting attribution methods by model, task, and language appear in Appendix Table~\ref{tab:guidelines_appendix}.

\section{Conclusion}

Faithfulness evaluation depends on the intervention operator, and this dependence has practical consequences. \ice{} operationalizes this insight through randomized baselines under multiple operators. Across our evaluation, operator gaps reach 44\,pp, while randomized baselines reveal that 18 of 56 English deletion configurations are anti-faithful. These patterns hold across 6 non-English languages and 2 attribution methods, and the methodology extends to step-level chain-of-thought evaluation (Appendix~\ref{app:cot}).

For practitioners, the recommendation is straightforward: report faithfulness under at least two operators, compare against random baselines, and treat disagreement between operators as genuine uncertainty about the explanation's quality.

\section*{Limitations}

\paragraph{Computational cost.} ICE's randomization tests ($M=50$) increase computation $50\times$ over single-point metrics. While this is tractable for the models evaluated here (1.5B--8B), scaling to larger LLMs (70B+) requires either subsampling or adaptive early stopping, which we have not yet validated.

\paragraph{Attribution method scope.} Our attention extraction averages across all layers and heads---layer-specific or head-specific analysis may yield finer-grained insights. We omit Integrated Gradients~\cite{sundararajan2017ig} for 7B+ LLMs due to memory constraints, though this method shows strong results on encoder models (Appendix~\ref{app:encoder}).

\paragraph{Multilingual coverage.} While our 6-language evaluation spans 4 scripts and 4 language families, languages with complex morphology beyond Turkish (e.g., Finnish, Japanese) and tonal languages (e.g., Vietnamese) remain untested. DeepSeek's tokenizer failure on Hindi and Chinese under retrieval infill highlights that operator applicability is model-dependent.

\paragraph{Behavioral vs.\ mechanistic faithfulness.} ICE evaluates \emph{behavioral} faithfulness---whether attributed tokens correlate with prediction changes---rather than \emph{mechanistic} faithfulness, which would require probing internal representations. Behavioral evaluation remains the most scalable approach for billion-parameter models but does not explain \emph{why} attributions are or are not faithful.

\paragraph{Extension to Chain-of-Thought.} ICE's operator-aware methodology generalizes to step-level CoT evaluation, where the unit of analysis changes from token spans to reasoning sentences. Appendix~\ref{app:cot} presents preliminary evidence that operator sensitivity persists at this level and that high accuracy does not imply faithful reasoning. We scope this paper to the token level, where model internals (attention, gradients) are available for attribution; a comprehensive step-level evaluation across frontier models is treated separately.

\paragraph{Retrieval infill limitations.} While retrieval infill mitigates OOD artifacts from deletion, replacement tokens drawn from the same distribution may preserve some task-relevant signal, potentially underestimating faithfulness. The GPT-2 Hindi case (retrieval exceeding deletion) suggests this concern is language- and model-dependent rather than systematic.

\section*{Reproducibility}

\noindent\fbox{\parbox{0.95\columnwidth}{\small
\textbf{Code}: \url{https://github.com/abhinaba/ice-faithfulness} \\
\textbf{Package}: \texttt{pip install ice-faithfulness} (\url{https://pypi.org/project/ice-faithfulness/}) \\
\textbf{Data}: All result JSON files across 12 experiment categories, each containing per-example scores, win rates, and bootstrap CIs (supplementary archive). \\
\textbf{Scripts}: \texttt{run\_ice\_llm.py} (English), \texttt{run\_ice\_multilingual.py} (6 languages), \texttt{run\_ice\_llm\_retrieval.py} (operator comparison). \\
\textbf{Runtime}: ${\sim}$15 min per model/dataset on RTX 4090 ($N{=}500$, $M{=}50$). Total: ${\sim}$200 GPU-hours. Pre-computed results included.
}}

\section*{Ethics Statement}

We caution against using faithfulness scores alone for high-stakes domains without domain validation. ``Faithful to model'' is not ``correct reasoning''---a model may faithfully rely on spurious correlations. Our experiments used ${\sim}$200 GPU-hours on RTX 4090/A100 hardware; we release pre-computed results to reduce replication cost.

\section*{Acknowledgments}

\paragraph{AI Assistance} We used AI assistants (Claude, Gemini, ChatGPT) for proofreading, editing, and verification of numerical consistency. All scientific contributions, experimental design, and writing are the authors' original work.

\bibliographystyle{acl_natbib}

\begin{thebibliography}{32}
\providecommand{\natexlab}[1]{#1}


\bibitem[{Amini et~al.(2025)Amini, Banaszak, Benoit, and others}]{liquidai2025lfm2}
Alexander Amini, Anna Banaszak, Harold Benoit, and others. 2025.
\newblock \href {https://arxiv.org/abs/2511.23404} {{LFM2} technical report}.
\newblock \emph{arXiv preprint arXiv:2511.23404}.

\bibitem[{Atanasova et~al.(2020)Atanasova, Simonsen, Lioma, and Augenstein}]{atanasova2020diagnostic}
Pepa Atanasova, Jakob~Grue Simonsen, Christina Lioma, and Isabelle Augenstein. 2020.
\newblock \href {https://doi.org/10.18653/v1/2020.emnlp-main.263} {A diagnostic study of explainability techniques for text classification}.
\newblock In \emph{Proceedings of the 2020 Conference on Empirical Methods in Natural Language Processing}, pages 3256--3274.

\bibitem[{Blard(2020)}]{blard2020allocine}
Th{\'e}ophile Blard. 2020.
\newblock French sentiment analysis with {BERT}.
\newblock
  \url{https://github.com/TheophileBlard/french-sentiment-analysis-with-bert}.

\bibitem[{Camburu et~al.(2018)Camburu, Rockt\"{a}schel, Lukasiewicz, and
  Blunsom}]{camburu2018esnli}
Oana-Maria Camburu, Tim Rockt\"{a}schel, Thomas Lukasiewicz, and Phil Blunsom.
  2018.
\newblock \href {https://proceedings.neurips.cc/paper_files/paper/2018/hash/4c7a167bb329bd92580a99ce422d6fa6-Abstract.html} {e-{SNLI}: Natural language inference with natural language explanations}.
\newblock In \emph{Proceedings of the 32nd International Conference on Neural
  Information Processing Systems}, NIPS'18, pages 9539--9549.

\bibitem[{DeepSeek-AI et~al.(2024)DeepSeek-AI, Bi, Chen, Chen, Chen, Dai,
  Deng, Ding, Dong, Du, Fu, Gao, Gao, Gao, Ge, Guan, Guo, Guo, Hao, Hao, He,
  Hu, Huang, Li, Li, Li, Li, Li, Liang, Lin, Liu, Liu, Liu, Liu, Liu, Liu, Lu,
  Lu, Luo, Ma, Nie, Pei, Piao, Qiu, Qu, Ren, Ren, Ruan, Sha, Shao, Song, Su,
  Sun, Sun, Tang, Wang, Wang, Wang, Wang, Wang, Wu, Wu, Xie, Xie, Xie, Xiong,
  Xu, Xu, Xu, Yang, You, Yu, Yu, Zhang, Zhang, Zhang, Zhang, Zhang, Zhang,
  Zhang, Zhang, Zhao, Zhao, Zhou, Zhou, Zhu, and Zou}]{deepseekllm}
DeepSeek-AI, Xiao Bi, Deli Chen, Guanting Chen, Shanhuang Chen, Damai Dai,
  Chengqi Deng, Honghui Ding, Kai Dong, Qiushi Du, Zhe Fu, Huazuo Gao, Kaige
  Gao, Wenjun Gao, Ruiqi Ge, Kang Guan, Daya Guo, Jianzhong Guo, and 69 others.
  2024.
\newblock \href {https://arxiv.org/abs/2401.02954} {{DeepSeek} {LLM}: Scaling
  open-source language models with longtermism}.
\newblock \emph{Preprint}, arXiv:2401.02954.

\bibitem[{Demirtas and Pechenizkiy(2013)}]{turkishsentiment2013}
Erkin Demirtas and Mykola Pechenizkiy. 2013.
\newblock \href {https://doi.org/10.1145/2502069.2502078} {Cross-lingual
  polarity detection with machine translation}.
\newblock In \emph{Proceedings of the Second International Workshop on Issues
  of Sentiment Discovery and Opinion Mining (WISDOM '13)}, pages 9:1--9:8. ACM.

\bibitem[{Devlin et~al.(2019)Devlin, Chang, Lee, and Toutanova}]{devlin2019bert}
Jacob Devlin, Ming-Wei Chang, Kenton Lee, and Kristina Toutanova. 2019.
\newblock \href {https://doi.org/10.18653/v1/N19-1423} {{BERT}: Pre-training of deep bidirectional transformers for language understanding}.
\newblock In \emph{Proceedings of NAACL-HLT 2019}, pages 4171--4186.

\bibitem[{DeYoung et~al.(2020)DeYoung, Jain, Rajani, Lehman, Xiong, Socher, and
  Wallace}]{deyoung2020eraser}
Jay DeYoung, Sarthak Jain, Nazneen~Fatema Rajani, Eric Lehman, Caiming Xiong,
  Richard Socher, and Byron~C. Wallace. 2020.
\newblock \href {https://aclanthology.org/2020.acl-main.408/} {{ERASER}: A
  Benchmark to Evaluate Rationalized {NLP} Models}.
\newblock In \emph{Proceedings of the 58th Annual Meeting of the Association
  for Computational Linguistics}, pages 4443--4458.

\bibitem[{Doddapaneni et~al.(2023)Doddapaneni, Aralikatte, Ramesh, Goyal,
  Khapra, Kunchukuttan, and Kumar}]{ai4bharat2020indicsentiment}
Sumanth Doddapaneni, Rahul Aralikatte, Gowtham Ramesh, Shreya Goyal, Mitesh~M.
  Khapra, Anoop Kunchukuttan, and Pratyush Kumar. 2023.
\newblock \href {https://doi.org/10.18653/v1/2023.acl-long.693} {Towards
  leaving no {I}ndic language behind: Building monolingual corpora, benchmark
  and models for {I}ndic languages}.
\newblock In \emph{Proceedings of the 61st Annual Meeting of the Association
  for Computational Linguistics (Volume 1: Long Papers)}, pages 12402--12426.
  Association for Computational Linguistics.

\bibitem[{Grattafiori et~al.(2024)Grattafiori, Dubey, Jauhri, Pandey, Kadian,
  Al-Dahle, Letman, Mathur, Schelten, Vaughan, Yang, Fan, Goyal, Hartshorn,
  Yang, Mitra, Sravankumar, Korenev, Hinsvark, Rao, Zhang, Rodriguez,
  Gregerson, Spataru, Roziere, Biron, Tang, Chern, Caucheteux, Nayak, Bi,
  Marra, McConnell, Keller, Touret, Wu, Wong, Ferrer, Nikolaidis, Allonsius,
  Song, Pintz, Livshits, Wyatt, Esiobu, Choudhary, Mahajan, Garcia-Olano,
  Perino, Hupkes, Lakomkin, AlBadawy, Lobanova, Dinan, Smith, Radenovic,
  Guzmán, Zhang, Synnaeve, Lee, Anderson, Thattai, Nail, Mialon, Pang,
  Cucurell, Nguyen, Korevaar, Xu, Touvron, Zarov, Ibarra, Kloumann, Misra,
  Evtimov, Zhang, Copet, Lee, Geffert, Vranes, Park, Mahadeokar, Shah, van~der
  Linde, Billock, Hong, Lee, Fu, Chi, Huang, Liu, Wang, Yu, Bitton, Spisak,
  Park, Rocca, Johnstun, Saxe, Jia, Alwala, Prasad, Upasani, Plawiak, Li,
  Heafield, Stone, El-Arini, Iyer, Malik, Chiu, Bhalla, Lakhotia,
  Rantala-Yeary, van~der Maaten, Chen, Tan, Jenkins, Martin, Madaan, Malo,
  Blecher, Landzaat, de~Oliveira, Muzzi, Pasupuleti, Singh, Paluri, Kardas,
  Tsimpoukelli, Oldham, Rita, Pavlova, Kambadur, Lewis, Si, Singh, Hassan,
  Goyal, Torabi, Bashlykov, Bogoychev, Chatterji, Zhang, Duchenne, Çelebi,
  Alrassy, Zhang, Li, Vasic, Weng, Bhargava, Dubal, Krishnan, Koura, Xu, He,
  Dong, Srinivasan, Ganapathy, Calderer, Cabral, Stojnic, Raileanu, Maheswari,
  Girdhar, Patel, Sauvestre, Polidoro, Sumbaly, Taylor, Silva, Hou, Wang,
  Hosseini, Chennabasappa, Singh, Bell, Kim, Edunov, Nie, Narang, Raparthy,
  Shen, Wan, Bhosale, Zhang, Vandenhende, Batra, Whitman, Sootla, Collot,
  Gururangan, Borodinsky, Herman, Fowler, Sheasha, Georgiou, Scialom,
  Speckbacher, Mihaylov, Xiao, Karn, Goswami, Gupta, Ramanathan, Kerkez,
  Gonguet, Do, Vogeti, Albiero, Petrovic, Chu, Xiong, Fu, Meers, Martinet,
  Wang, Wang, Tan, Xia, Xie, Jia, Wang, Goldschlag, Gaur, Babaei, Wen, Song,
  Zhang, Li, Mao, Coudert, Yan, Chen, Papakipos, Singh, Srivastava, Jain,
  Kelsey, Shajnfeld, Gangidi, Victoria, Goldstand, Menon, Sharma, Boesenberg,
  Baevski, Feinstein, Kallet, Sangani, Teo, Yunus, Lupu, Alvarado, Caples, Gu,
  Ho, Poulton, Ryan, Ramchandani, Dong, Franco, Goyal, Saraf, Chowdhury,
  Gabriel, Bharambe, Eisenman, Yazdan, James, Maurer, Leonhardi, Huang, Loyd,
  Paola, Paranjape, Liu, Wu, Ni, Hancock, Wasti, Spence, Stojkovic, Gamido,
  Montalvo, Parker, Burton, Mejia, Liu, Wang, Kim, Zhou, Hu, Chu, Cai, Tindal,
  Feichtenhofer, Gao, Civin, Beaty, Kreymer, Li, Adkins, Xu, Testuggine, David,
  Parikh, Liskovich, Foss, Wang, Le, Holland, Dowling, Jamil, Montgomery,
  Presani, Hahn, Wood, Le, Brinkman, Arcaute, Dunbar, Smothers, Sun, Kreuk,
  Tian, Kokkinos, Ozgenel, Caggioni, Kanayet, Seide, Florez, Schwarz, Badeer,
  Swee, Halpern, Herman, Sizov, Guangyi, Zhang, Lakshminarayanan, Inan,
  Shojanazeri, Zou, Wang, Zha, Habeeb, Rudolph, Suk, Aspegren, Goldman, Zhan,
  Damlaj, Molybog, Tufanov, Leontiadis, Veliche, Gat, Weissman, Geboski, Kohli,
  Lam, Asher, Gaya, Marcus, Tang, Chan, Zhen, Reizenstein, Teboul, Zhong, Jin,
  Yang, Cummings, Carvill, Shepard, McPhie, Torres, Ginsburg, Wang, Wu, U,
  Saxena, Khandelwal, Zand, Matosich, Veeraraghavan, Michelena, Li, Jagadeesh,
  Huang, Chawla, Huang, Chen, Garg, A, Silva, Bell, Zhang, Guo, Yu, Moshkovich,
  Wehrstedt, Khabsa, Avalani, Bhatt, Mankus, Hasson, Lennie, Reso, Groshev,
  Naumov, Lathi, Keneally, Liu, Seltzer, Valko, Restrepo, Patel, Vyatskov,
  Samvelyan, Clark, Macey, Wang, Hermoso, Metanat, Rastegari, Bansal,
  Santhanam, Parks, White, Bawa, Singhal, Egebo, Usunier, Mehta, Laptev, Dong,
  Cheng, Chernoguz, Hart, Salpekar, Kalinli, Kent, Parekh, Saab, Balaji,
  Rittner, Bontrager, Roux, Dollar, Zvyagina, Ratanchandani, Yuvraj, Liang,
  Alao, Rodriguez, Ayub, Murthy, Nayani, Mitra, Parthasarathy, Li, Hogan,
  Battey, Wang, Howes, Rinott, Mehta, Siby, Bondu, Datta, Chugh, Hunt, Dhillon,
  Sidorov, Pan, Mahajan, Verma, Yamamoto, Ramaswamy, Lindsay, Lindsay, Feng,
  Lin, Zha, Patil, Shankar, Zhang, Zhang, Wang, Agarwal, Sajuyigbe, Chintala,
  Max, Chen, Kehoe, Satterfield, Govindaprasad, Gupta, Deng, Cho, Virk,
  Subramanian, Choudhury, Goldman, Remez, Glaser, Best, Koehler, Robinson, Li,
  Zhang, Matthews, Chou, Shaked, Vontimitta, Ajayi, Montanez, Mohan, Kumar,
  Mangla, Ionescu, Poenaru, Mihailescu, Ivanov, Li, Wang, Jiang, Bouaziz,
  Constable, Tang, Wu, Wang, Wu, Gao, Kleinman, Chen, Hu, Jia, Qi, Li, Zhang,
  Zhang, Adi, Nam, Yu, Wang, Zhao, Hao, Qian, Li, He, Rait, DeVito, Rosnbrick,
  Wen, Yang, Zhao, and Ma}]{llama3model}
Aaron Grattafiori, Abhimanyu Dubey, Abhinav Jauhri, Abhinav Pandey, Abhishek
  Kadian, Ahmad Al-Dahle, Aiesha Letman, Akhil Mathur, Alan Schelten, Alex
  Vaughan, Amy Yang, Angela Fan, Anirudh Goyal, Anthony Hartshorn, Aobo Yang,
  Archi Mitra, Archie Sravankumar, Artem Korenev, Arthur Hinsvark, and 542
  others. 2024.
\newblock \href {https://arxiv.org/abs/2407.21783} {The llama 3 herd of
  models}.
\newblock \emph{Preprint}, arXiv:2407.21783.

\bibitem[{Hase et~al.(2021)Hase, Xie, and Bansal}]{hase2021ood}
Peter Hase, Harry Xie, and Mohit Bansal. 2021.
\newblock \href {https://proceedings.neurips.cc/paper/2021/hash/1def1713ebf17722cbe300cfc1c88558-Abstract.html} {The
  out-of-distribution problem in explainability and search methods for feature
  importance explanations}.
\newblock In \emph{Proceedings of the 35th International Conference on Neural
  Information Processing Systems}, NeurIPS '21, pages 3650--3666.

\bibitem[{Jacovi and Goldberg(2020)}]{jacovi2020faithfully}
Alon Jacovi and Yoav Goldberg. 2020.
\newblock \href {https://doi.org/10.18653/v1/2020.acl-main.386} {Towards
  faithfully interpretable {NLP} systems: How should we define and evaluate
  faithfulness?}
\newblock In \emph{Proceedings of the 58th Annual Meeting of the Association
  for Computational Linguistics}, pages 4198--4205. Association for
  Computational Linguistics.

\bibitem[{Jiang et~al.(2023)Jiang, Sablayrolles, Mensch, Bamford, Chaplot,
  de~las Casas, Bressand, Lengyel, Lample, Saulnier, Lavaud, Lachaux, Stock,
  Scao, Lavril, Wang, Lacroix, and Sayed}]{mistral7b}
Albert~Q. Jiang, Alexandre Sablayrolles, Arthur Mensch, Chris Bamford,
  Devendra~Singh Chaplot, Diego de~las Casas, Florian Bressand, Gianna Lengyel,
  Guillaume Lample, Lucile Saulnier, Lélio~Renard Lavaud, Marie-Anne Lachaux,
  Pierre Stock, Teven~Le Scao, Thibaut Lavril, Thomas Wang, Timothée Lacroix,
  and William~El Sayed. 2023.
\newblock \href {https://arxiv.org/abs/2310.06825} {Mistral {7B}}.
\newblock \emph{Preprint}, arXiv:2310.06825.

\bibitem[{Kamp et~al.(2023)Kamp, Beinborn, and Fokkens}]{kamp2023dynamic}
Jonathan Kamp, Lisa Beinborn, and Antske Fokkens. 2023.
\newblock \href {https://doi.org/10.18653/v1/2023.emnlp-main.379} {Dynamic
  top-k estimation consolidates disagreement between feature attribution
  methods}.
\newblock In \emph{Proceedings of the 2023 Conference on Empirical Methods in
  Natural Language Processing}, pages 6190--6197. Association for Computational
  Linguistics.

\bibitem[{Lanham et~al.(2023)Lanham, Chen, Radhakrishnan, Steiner, Denison, Hernandez, Li, Durmus, Hubinger, Kerr, and others}]{lanham2023measuring}
Tamera Lanham, Anna Chen, Ansh Radhakrishnan, Benoit Steiner, Carson Denison, Danny Hernandez, Dustin Li, Esin Durmus, Evan Hubinger, Jackson Kernion, and others. 2023.
\newblock \href {https://arxiv.org/abs/2307.13702} {Measuring faithfulness in chain-of-thought reasoning}.
\newblock \emph{arXiv preprint arXiv:2307.13702}.

\bibitem[{Maas et~al.(2011)Maas, Daly, Pham, Huang, Ng, and
  Potts}]{maas2011learning}
Andrew~L. Maas, Raymond~E. Daly, Peter~T. Pham, Dan Huang, Andrew~Y. Ng, and
  Christopher Potts. 2011.
\newblock \href {https://aclanthology.org/P11-1015/} {Learning word vectors for
  sentiment analysis}.
\newblock In \emph{Proceedings of the 49th Annual Meeting of the Association
  for Computational Linguistics: Human Language Technologies}, pages 142--150.
  Association for Computational Linguistics.

\bibitem[{Madsen et~al.(2024)Madsen, Chandar, and Reddy}]{madsen2024faithful}
Andreas Madsen, Sarath Chandar, and Siva Reddy. 2024.
\newblock \href {https://doi.org/10.18653/v1/2024.findings-acl.19} {Are
  Self-Explanations from Large Language Models Faithful?}
\newblock In \emph{Findings of the Association for Computational Linguistics:
  ACL 2024}, pages 295--337. Association for Computational Linguistics.

\bibitem[{Meta(2024)}]{llama32model}
Meta. 2024.
\newblock \href {https://ai.meta.com/blog/llama-3-2-connect-2024-vision-edge-mobile-devices/} {Llama 3.2: Revolutionizing edge {AI} and vision with open, customizable models}.
\newblock Meta AI Blog, September 2024.

\bibitem[{Nabil et~al.(2015)Nabil, Aly, and Atiya}]{arabicsentiment2015}
Mahmoud Nabil, Mohamed Aly, and Amir~F. Atiya. 2015.
\newblock \href {https://doi.org/10.18653/v1/D15-1299} {{ASTD}: {A}rabic
  sentiment tweets dataset}.
\newblock In \emph{Proceedings of the 2015 Conference on Empirical Methods in
  Natural Language Processing}, pages 2515--2519. Association for Computational
  Linguistics.

\bibitem[{{NLLB Team} et~al.(2022){NLLB Team}, Costa-juss{\`a}, and others}]{nllb2022}
{NLLB Team}, Marta~R. Costa-juss{\`a}, and others. 2022.
\newblock \href {https://arxiv.org/abs/2207.04672} {No language left behind: Scaling human-centered machine translation}.
\newblock \emph{Preprint}, arXiv:2207.04672.

\bibitem[{Radford et~al.(2019)Radford, Wu, Child, Luan, Amodei, and
  Sutskever}]{radford2019language}
Alec Radford, Jeffrey Wu, Rewon Child, David Luan, Dario Amodei, and Ilya
  Sutskever. 2019.
\newblock \href
  {https://cdn.openai.com/better-language-models/language_models_are_unsupervised_multitask_learners.pdf}
  {Language models are unsupervised multitask learners}.
\newblock Technical report, OpenAI.

\bibitem[{Ribeiro et~al.(2016)Ribeiro, Singh, and Guestrin}]{ribeiro2016lime}
Marco~Tulio Ribeiro, Sameer Singh, and Carlos Guestrin. 2016.
\newblock \href {https://doi.org/10.1145/2939672.2939778} {``Why should {I} trust you?'' Explaining the predictions of any classifier}.
\newblock In \emph{Proceedings of the 22nd ACM SIGKDD International Conference on Knowledge Discovery and Data Mining}, pages 1135--1144.

\bibitem[{Rust et~al.(2021)Rust, Pfeiffer, Vulić, Ruder, and Gurevych}]{rust2021good}
Phillip Rust, Jonas Pfeiffer, Ivan Vulić, Sebastian Ruder, and Iryna Gurevych. 2021.
\newblock \href {https://aclanthology.org/2021.acl-long.243/} {How good is your tokenizer? {O}n the monolingual performance of multilingual language models}.
\newblock In \emph{Proceedings of the 59th Annual Meeting of the ACL}, pages 3118--3135.

\bibitem[{Socher et~al.(2013)Socher, Perelygin, Wu, Chuang, Manning, Ng, and
  Potts}]{socher2013recursive}
Richard Socher, Alex Perelygin, Jean~Y. Wu, Jason Chuang, Christopher~D. Manning,
  Andrew~Y. Ng, and Christopher Potts. 2013.
\newblock \href {https://aclanthology.org/D13-1170/} {Recursive deep models for
  semantic compositionality over a sentiment treebank}.
\newblock In \emph{Proceedings of the 2013 Conference on Empirical Methods in
  Natural Language Processing}, pages 1631--1642. Association for Computational
  Linguistics.

\bibitem[{Sun et~al.(2025)Sun, Atanasova, and Augenstein}]{sun2025unified}
Jingyi Sun, Pepa Atanasova, and Isabelle Augenstein. 2025.
\newblock \href {https://doi.org/10.18653/v1/2025.naacl-long.530} {Evaluating
  input feature explanations through a unified diagnostic evaluation
  framework}.
\newblock In \emph{Proceedings of the 2025 Conference of the Nations of the
  Americas Chapter of the Association for Computational Linguistics: Human
  Language Technologies (Volume 1: Long Papers)}, pages 10559--10577.
  Association for Computational Linguistics.

\bibitem[{Sundararajan et~al.(2017)Sundararajan, Taly, and Yan}]{sundararajan2017ig}
Mukund Sundararajan, Ankur Taly, and Qiqi Yan. 2017.
\newblock \href {https://proceedings.mlr.press/v70/sundararajan17a.html} {Axiomatic attribution for deep networks}.
\newblock In \emph{Proceedings of the 34th International Conference on Machine Learning (ICML)}, PMLR 70, pages 3319--3328.

\bibitem[{Tan and Zhang(2008)}]{tan2008chnsenticorp}
Songbo Tan and Jin Zhang. 2008.
\newblock \href {https://doi.org/10.1016/j.eswa.2007.05.028} {An empirical
  study of sentiment analysis for chinese documents}.
\newblock \emph{Expert Systems with Applications}, 34(4):2622--2629.

\bibitem[{Turpin et~al.(2023)Turpin, Michael, Perez, and Bowman}]{turpin2024language}
Miles Turpin, Julian Michael, Ethan Perez, and Samuel~R. Bowman. 2023.
\newblock \href {https://proceedings.neurips.cc/paper_files/paper/2023/hash/ed3fea9033a80fea1376299c7f25547f-Abstract-Conference.html} {Language models don't always say what they think: Unfaithful explanations in chain-of-thought prompting}.
\newblock In \emph{Advances in Neural Information Processing Systems}, volume~36, pages 74952--74965.

\bibitem[{Wojatzki et~al.(2017)Wojatzki, Ruppert, Holschneider, Zesch, and
  Biemann}]{wojatzki2017germeval}
Michael Wojatzki, Eugen Ruppert, Sarah Holschneider, Torsten Zesch, and Chris
  Biemann. 2017.
\newblock \href {https://ids-pub.bsz-bw.de/frontdoor/deliver/index/docId/6916/file/Wojatzki_etal._GermEval_2017_2017.pdf} {Germeval 2017: Shared task on aspect-based sentiment in social media customer feedback}.
\newblock In \emph{Proceedings of the GermEval 2017 – Shared Task on
  Aspect-based Sentiment in Social Media Customer Feedback}, pages 1--12.

\bibitem[{Yang et~al.(2024)Yang, Yang, Zhang, Hui, Zheng, Yu, and others}]{qwen25}
An~Yang, Baosong Yang, Beichen Zhang, Binyuan Hui, Bo~Zheng, Bowen~Yu, and others. 2024.
\newblock \href {https://arxiv.org/abs/2412.15115} {Qwen2.5 technical report}.
\newblock \emph{Preprint}, arXiv:2412.15115.

\bibitem[{Zaman and Srivastava(2025)}]{zaman2025causal}
Kerem Zaman and Shashank Srivastava. 2025.
\newblock \href {https://doi.org/10.18653/v1/2025.emnlp-main.1496} {A causal
  lens for evaluating faithfulness metrics}.
\newblock In \emph{Proceedings of the 2025 Conference on Empirical Methods in
  Natural Language Processing}, pages 29425--29449. Association for
  Computational Linguistics.

\bibitem[{Zhang et~al.(2015)Zhang, Zhao, and LeCun}]{zhang2015character}
Xiang Zhang, Junbo Zhao, and Yann LeCun. 2015.
\newblock \href {https://proceedings.neurips.cc/paper/2015/hash/250cf8b51c773f3f8dc8b4be867a9a02-Abstract.html} {Character-level
  convolutional networks for text classification}.
\newblock In \emph{Proceedings of the 29th International Conference on Neural
  Information Processing Systems - Volume 1}, NIPS'15, pages 649--657.

\bibitem[{Zheng et~al.(2025)Zheng, Shirani, Chen, Lin, Cheng, Guo, and
  Luo}]{zheng2025ffidelity}
Xu~Zheng, Farhad Shirani, Zhuomin Chen, Chaohao Lin, Wei Cheng, Wenbo Guo, and
  Dongsheng Luo. 2025.
\newblock \href {https://iclr.cc/virtual/2025/poster/29329} {F-Fidelity: A
  robust framework for faithfulness evaluation of explainable {AI}}.
\newblock In \emph{The Thirteenth International Conference on Learning
  Representations}.

\end{thebibliography}

\clearpage
\appendix

\section*{\centering --- Appendix ---}

\section{Operator Ablation}
\label{app:operator_ablation}

Table~\ref{tab:operator_ablation} compares operator types on GPT-2/SST-2.

\begin{table}[H]
\centering
\small
\begin{tabular}{lcc}
\toprule
\textbf{Operator} & \textbf{Attn WR} & \textbf{Grad WR} \\
\midrule
Deletion & 59.8\% & 52.8\% \\
Mask-UNK & 12.9\% & 22.0\% \\
Mask-PAD & 13.4\% & 21.6\% \\
\bottomrule
\end{tabular}
\caption{Operator ablation on GPT-2/SST-2 (N=100). Masking produces degenerate outputs for autoregressive LLMs, with win rates 4--5$\times$ lower than deletion.}
\label{tab:operator_ablation}
\end{table}

\section{Models and Datasets}
\label{app:models_datasets}

Table~\ref{tab:models} lists the evaluated models. Table~\ref{tab:versions} provides pinned dataset revisions.

\begin{table}[H]
\centering
\small
\begin{tabular}{llcp{3.0cm}}
\toprule
\textbf{Model} & \textbf{Size} & \textbf{Type} & \textbf{Description} \\
\midrule
GPT-2 & 1.5B & Base & Baseline autoregressive \cite{radford2019language} \\
LFM2 & 2.6B & Base & Efficient architecture \cite{liquidai2025lfm2} \\
Llama 3.2 & 3B & Inst & Small instruction-tuned \cite{llama32model} \\
Llama 3.1 & 8B & Base & General reasoning \cite{llama3model} \\
Qwen 2.5 & 7B & Inst & Multilingual focus \cite{qwen25} \\
Mistral & 7B & Inst & Efficient instruction \cite{mistral7b} \\
DeepSeek & 7B & Chat & Multilingual/Chat \cite{deepseekllm} \\
\bottomrule
\end{tabular}
\caption{Evaluated LLMs spanning diverse sizes and capabilities.}
\label{tab:models}
\end{table}

English datasets: SST-2 (binary sentiment, short text) \cite{socher2013recursive}, IMDB (binary sentiment, long text) \cite{maas2011learning}, e-SNLI (NLI with human rationales) \cite{camburu2018esnli}, AG News (4-class topic classification) \cite{zhang2015character}.

Multilingual: French (Allocine \cite{blard2020allocine}), German (GermEval 2017 \cite{wojatzki2017germeval}), Hindi (IndicSentiment \cite{ai4bharat2020indicsentiment}), Chinese (ChnSentiCorp \cite{tan2008chnsenticorp}), Turkish (Turkish Sentiment \cite{turkishsentiment2013}), Arabic (Arabic Sentiment \cite{arabicsentiment2015}).

\begin{table}[H]
\centering
\footnotesize
\begin{tabular}{ll}
\hline
\textbf{Dataset} & \textbf{Revision SHA} \\
\hline
glue (SST-2) & \texttt{bcdcba79d07bc86...} \\
imdb & \texttt{e6281661ce1c48d...} \\
esnli & \texttt{a160e6a02bbb8d8...} \\
ag\_news & \texttt{eb185aade064a81...} \\
\hline
\end{tabular}
\caption{Pinned dataset revisions for ICEBench.}
\label{tab:versions}
\end{table}

\section{Statistical Framework Details}
\label{app:stats}

Bootstrap 95\% confidence intervals are computed over $B=200$ resamples:
\begin{equation}
\text{CI}_{95\%} = [\text{NSR}_{2.5\%}^*, \text{NSR}_{97.5\%}^*]
\end{equation}

We apply Benjamini-Hochberg FDR correction at $\alpha = 0.10$ when evaluating multiple examples. We use magnitude thresholds 0.2 (small), 0.5 (medium), 0.8 (large) for descriptive interpretation. Negative $d_{\text{null}}$ indicates anti-faithfulness.

\section{K-Sensitivity Analysis}
\label{app:ksweep}

We evaluate faithfulness sensitivity to rationale length $k \in \{0.1, 0.2, 0.3, 0.4, 0.5\}$ on multilingual data (French/German) with GPT-2 and LFM2-2.6B.

\begin{table}[H]
\centering
\footnotesize
\setlength{\tabcolsep}{3pt}
\begin{tabular}{@{}llccccc@{}}
\hline
\textbf{Model} & \textbf{Extr.} & \textbf{0.1} & \textbf{0.2} & \textbf{0.3} & \textbf{0.4} & \textbf{0.5} \\
\hline
\multicolumn{7}{c}{\textit{German Win Rate (\%)}} \\
\hline
GPT-2 & Attn & 47.2 & 52.7 & 57.0 & 61.7 & \textbf{66.6} \\
GPT-2 & Grad & \textbf{54.5} & 46.2 & 49.7 & 47.1 & 42.0 \\
LFM2 & Attn & 42.2 & 52.3 & 46.1 & 49.3 & \textbf{55.2} \\
LFM2 & Grad & 50.9 & 47.0 & 51.1 & 52.1 & \textbf{55.9} \\
\hline
\multicolumn{7}{c}{\textit{French Win Rate (\%)}} \\
\hline
GPT-2 & Attn & 48.1 & 49.0 & 48.2 & 47.6 & 47.7 \\
GPT-2 & Grad & 48.7 & 49.6 & 49.8 & 49.2 & \textbf{50.2} \\
LFM2 & Attn & 76.6 & 64.0 & 66.2 & 77.1 & \textbf{79.1} \\
LFM2 & Grad & 60.2 & 66.9 & 71.8 & 75.3 & \textbf{78.1} \\
\hline
\end{tabular}
\caption{K-sensitivity: win rate (\%) by rationale length $k$. Bold = best per config. GPT-2 German gradient is non-monotonic (peaks at $k$=0.1), while attention increases with $k$.}
\label{tab:ksweep}
\end{table}

\section{Prompt Sensitivity Analysis}
\label{app:prompt_sensitivity}

\begin{table}[H]
\centering
\footnotesize
\setlength{\tabcolsep}{2.5pt}
\begin{tabular}{@{}llcccccc@{}}
\toprule
\textbf{Data} & \textbf{Prompt} & \textbf{Acc} & \textbf{Conf} & \textbf{Attn} & \textbf{Grad} & \textbf{$\Delta$} \\
\midrule
SST-2 & v1 (standard) & 52\% & 68.8 & \textbf{72.7} & \textbf{62.0} & +10.7 \\
& v2 (minimal) & 59\% & 65.2 & 43.4 & 55.2 & -11.8 \\
& v3 (question) & 72\% & 56.3 & 47.3 & 46.3 & +1.0 \\
& v4 (completion) & 64\% & 63.9 & 67.5 & 64.6 & +2.9 \\
& v5 (quoted) & 74\% & 57.0 & 60.3 & 50.3 & +10.0 \\
\midrule
IMDB & v1 (standard) & 42\% & 57.9 & 60.2 & 58.1 & +2.1 \\
& v2 (rating) & 2\% & 63.1 & 34.8$^\dagger$ & 33.2$^\dagger$ & +1.6 \\
& v3 (yes/no) & 20\% & 61.3 & \textbf{75.4} & \textbf{74.8} & +0.6 \\
\midrule
AG News & v1 (standard) & 47\% & 69.6 & \textbf{71.1} & 48.8 & +22.3 \\
& v2 (minimal) & 65\% & 79.2 & 62.8 & \textbf{67.9} & -5.1 \\
& v3 (question) & 73\% & 68.2 & 58.9 & 47.3 & +11.6 \\
\midrule
e-SNLI & v1 (standard) & 31\% & 77.0 & 63.7 & 14.7$^\dagger$ & +49.0 \\
& v2 (verb) & 31\% & 88.9 & \textbf{95.5} & 24.0$^\dagger$ & +71.5 \\
& v3 (T/F/U) & 34\% & 56.4 & 65.9 & \textbf{68.6} & -2.7 \\
\bottomrule
\end{tabular}
\caption{Prompt sensitivity analysis (GPT-2). Win rates (\%) for attention (Attn) and gradient (Grad). $\Delta$ = Attn $-$ Grad. $^\dagger$Anti-faithful ($<$50\%).}
\label{tab:prompt_sensitivity}
\end{table}

\begin{figure}[H]
\centering
\includegraphics[width=\columnwidth]{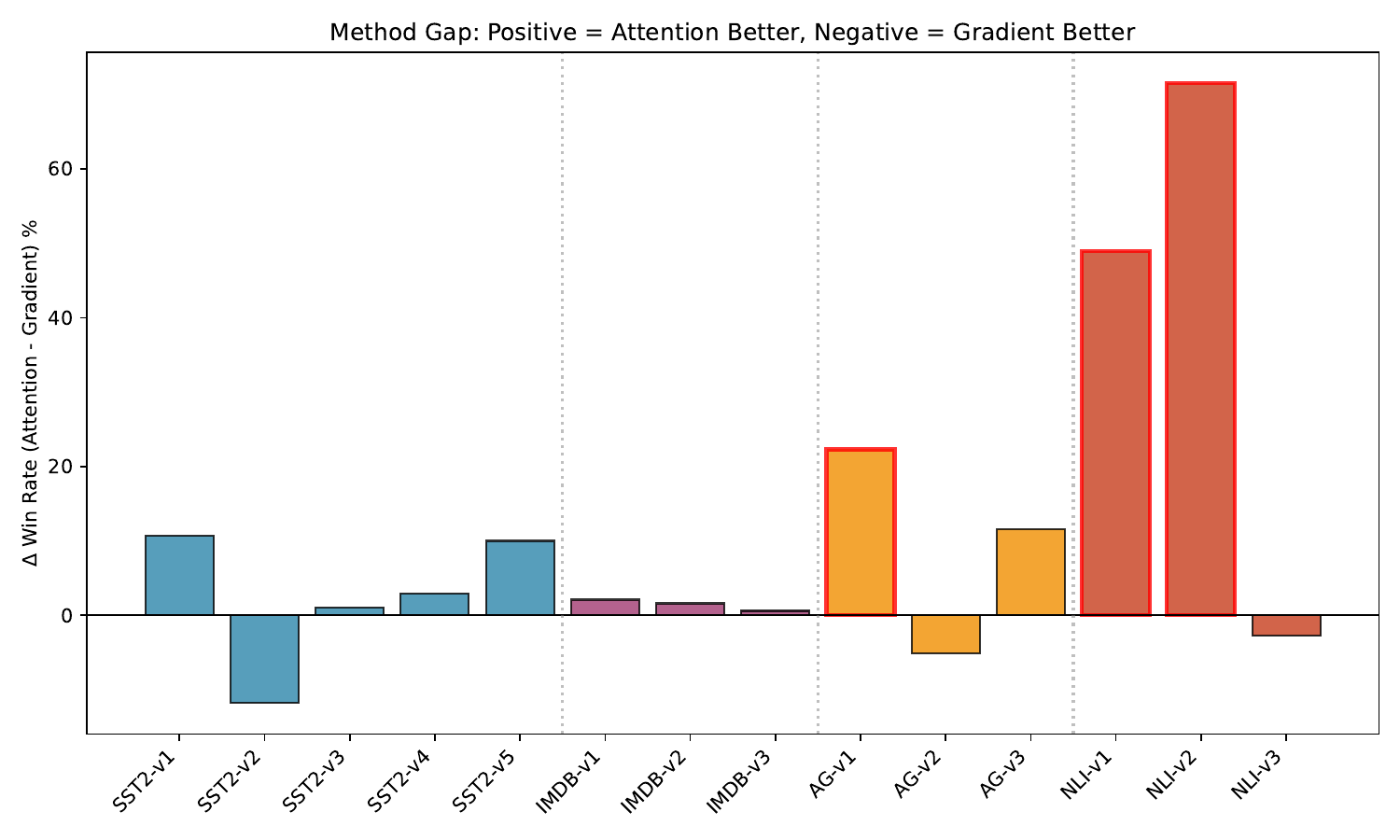}
\caption{Method gap ($\Delta$ = Attention $-$ Gradient win rate) by prompt variant.}
\label{fig:method_gap}
\end{figure}

\begin{figure}[H]
\centering
\includegraphics[width=\columnwidth]{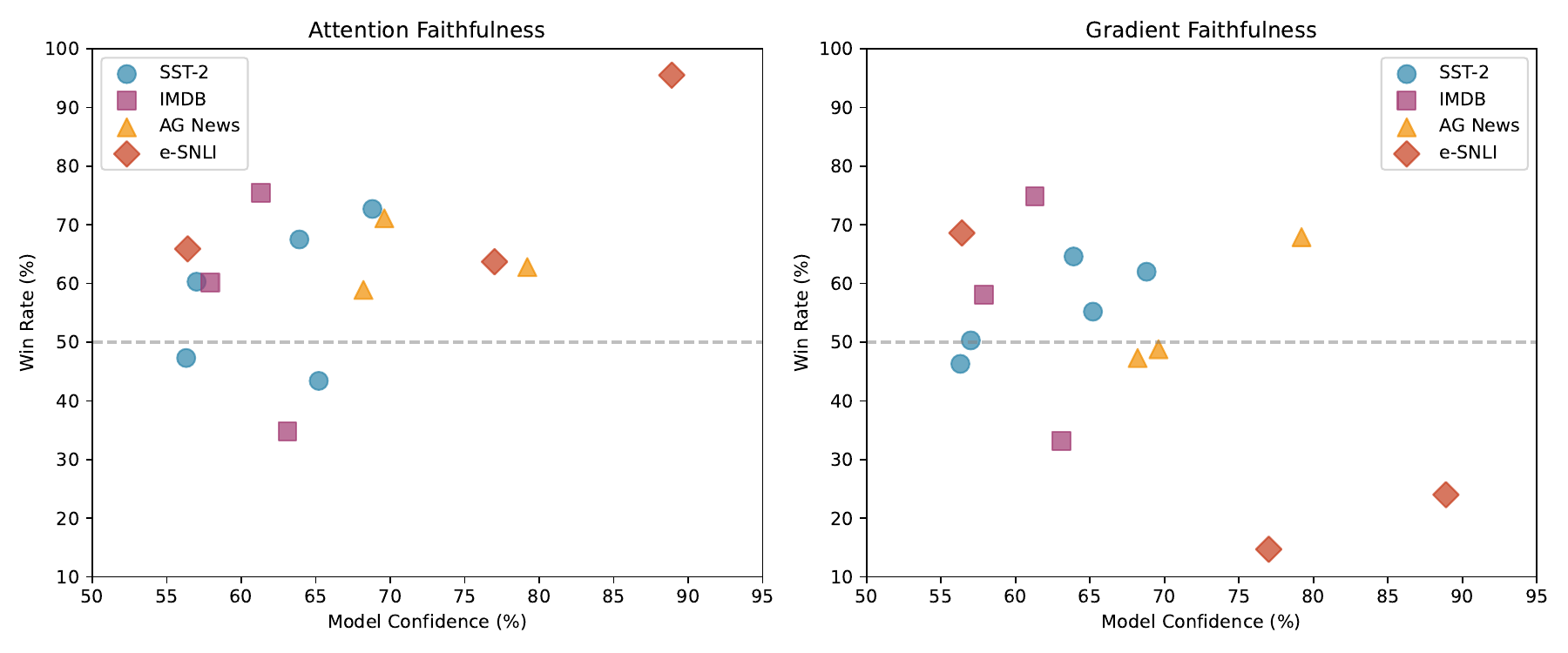}
\caption{Faithfulness vs.\ model confidence across prompt variants.}
\label{fig:faith_conf}
\end{figure}

\begin{figure}[H]
\centering
\includegraphics[width=0.85\columnwidth]{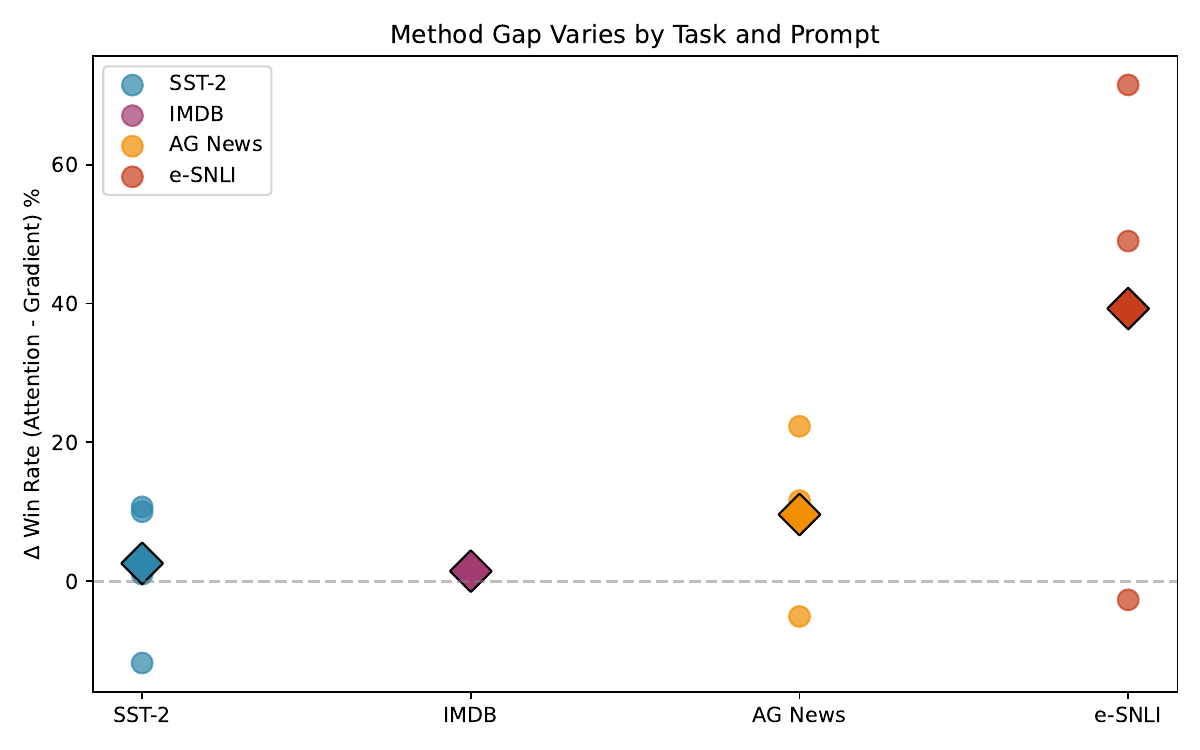}
\caption{Method gap distribution by task.}
\label{fig:gap_by_task}
\end{figure}

\section{Language-Specific Patterns}
\label{app:language_patterns}

\textbf{German}: High variance across models (34--83\%). Qwen is the clear outlier at 83\% attention, the highest multilingual result. We speculate that compound-word handling in Qwen's tokenizer may play a role, though we cannot isolate this factor.

\textbf{Chinese}: Moderate results (50--69\% deletion). LFM2 gradient leads (69\%), but drops under retrieval infill, suggesting operator sensitivity extends to character-level tokenization.

\textbf{French}: The most polarized language. GPT-2 is anti-faithful (15--16\%) while Llama 3.1 excels (81\% attention).

\textbf{Hindi}: GPT-2 performs surprisingly well (65--66\%) despite 8.1$\times$ token expansion. Under retrieval infill, GPT-2 reaches 68.8\% (exceeding deletion), while Llama 3.1 drops to 35.1\%.

\textbf{Turkish}: Qwen leads attention (79.3\%) while Mistral leads gradient (71.5\%), the largest method divergence for any language. GPT-2 is anti-faithful (29.7\%).

\textbf{Arabic}: Mistral attention achieves 72.6\% but its gradient drops to 39.9\%, the largest attention-gradient gap across languages.

\section{Multilingual Detailed Results}
\label{app:multilingual}

\begin{table}[H]
\centering
\footnotesize
\setlength{\tabcolsep}{3pt}
\begin{tabular}{@{}lcccccc@{}}
\toprule
\textbf{Model} & \textbf{FR} & \textbf{DE} & \textbf{HI} & \textbf{ZH} & \textbf{TR} & \textbf{AR} \\
\midrule
\multicolumn{7}{c}{\textit{Deletion --- Attention Win Rate (\%)}} \\
\midrule
GPT-2 & 15.8 & 49.1 & 65.4 & 57.5 & 29.7 & \textbf{62.4} \\
Llama 3.1-8B & 80.8 & 39.1 & 44.6 & 49.1 & 73.0 & 63.4 \\
Llama 3.2-3B & 44.9 & 49.0 & 53.7 & 58.3 & 59.2 & 37.1 \\
Qwen 2.5-7B & 62.6 & \textbf{82.7} & 50.2 & 60.6 & \textbf{79.3} & 57.0 \\
Mistral 7B & 53.5 & 65.6 & 53.1 & 60.2 & 57.2 & \textbf{72.6} \\
DeepSeek 7B & \textbf{67.2} & 43.2 & \textbf{65.4} & 56.1 & 49.0 & $\dagger$ \\
LFM2-2.6B & 59.6 & 48.6 & 62.3 & 52.7 & 73.7 & 69.0 \\
\midrule
\multicolumn{7}{c}{\textit{Retrieval Infill --- Attention Win Rate (\%)}} \\
\midrule
GPT-2 & 23.9 & 52.2 & \textbf{68.8} & 50.0 & 43.2 & 71.7 \\
Llama 3.1-8B & \textbf{81.8} & 35.1 & 35.1 & 37.3 & 70.3 & 44.8 \\
Llama 3.2-3B & 39.7 & 46.2 & 52.3 & 52.9 & 54.8 & 23.5 \\
Qwen 2.5-7B & 62.3 & 70.1 & 51.0 & 56.2 & \textbf{81.6} & 48.5 \\
Mistral 7B & 51.0 & 55.5 & 51.3 & 50.0 & 53.1 & \textbf{62.9} \\
DeepSeek 7B & 54.4 & 12.0 & $\dagger$ & $\dagger$ & 50.7 & 80.0 \\
LFM2-2.6B & 65.1 & 51.6 & 66.1 & \textbf{58.3} & 65.6 & 71.1 \\
\bottomrule
\end{tabular}
\caption{Multilingual attention win rates (\%) under both operators. $\dagger$\,=\,no valid output (Arabic: model limitation; HI/ZH retrieval: tokenizer failure on DeepSeek).}
\label{tab:multilingual}
\end{table}

\begin{table}[H]
\centering
\small
\begin{tabular}{lcccc}
\toprule
\textbf{Model} & \textbf{FR} & \textbf{DE} & \textbf{HI} & \textbf{ZH} \\
\midrule
GPT-2 & 1.8$\times$ & 2.0$\times$ & 8.1$\times$ & 10.7$\times$ \\
Llama 3.x & 1.3$\times$ & 1.4$\times$ & 2.5$\times$ & 4.2$\times$ \\
Mistral & 1.4$\times$ & 1.5$\times$ & 5.0$\times$ & 5.7$\times$ \\
\bottomrule
\end{tabular}
\caption{Token expansion ratios by language ($\times$ = tokens per character vs.\ English).}
\label{tab:tokenization}
\end{table}

\section{Encoder Validation Results}
\label{app:encoder}

Table~\ref{tab:encoder_results} reports ICE evaluation on BERT-base-uncased~\cite{devlin2019bert} across five ERASER datasets, comparing LIME~\cite{ribeiro2016lime}, Integrated Gradients~\cite{sundararajan2017ig}, attention, and gradient extractors.

\begin{table}[H]
\centering
\small
\begin{tabular}{lcccc}
\toprule
\textbf{Extractor} & \textbf{Suf.} & \textbf{Sig.\ Rate} & \textbf{AUC-Suf} \\
\midrule
\multicolumn{4}{c}{\textit{SST-2 (500 examples)}} \\
\midrule
LIME & \textbf{0.617} & \textbf{11.4\%} & \textbf{0.302} \\
Integrated Gradients & 0.492 & 0\% & 0.256 \\
Attention & 0.398 & 0\% & 0.250 \\
Gradient & 0.394 & 0\% & 0.253 \\
\midrule
\multicolumn{4}{c}{\textit{IMDB (500 examples)}} \\
\midrule
Gradient & 0.519 & \textbf{57.4\%} & 0.345 \\
Attention & 0.385 & 33.2\% & 0.346 \\
LIME & 0.182 & 0\% & 0.284 \\
Integrated Gradients & 0.149 & 0\% & 0.326 \\
\midrule
\multicolumn{4}{c}{\textit{e-SNLI (417 examples)$^\dagger$}} \\
\midrule
LIME & \textbf{0.450} & 0\% & 0.195 \\
Integrated Gradients & 0.406 & 0\% & 0.190 \\
Gradient & 0.383 & 0\% & 0.194 \\
Attention & 0.352 & 0\% & 0.152 \\
\midrule
\multicolumn{4}{c}{\textit{BoolQ (500 examples)}} \\
\midrule
Gradient & 0.071 & 0\% & 0.062 \\
Integrated Gradients & 0.070 & 0\% & 0.056 \\
Attention & 0.066 & 0\% & 0.055 \\
LIME & 0.058 & 0\% & 0.046 \\
\midrule
\multicolumn{4}{c}{\textit{MultiRC (500 examples)}} \\
\midrule
Attention & \textbf{0.103} & 0\% & \textbf{0.098} \\
Gradient & 0.079 & 0\% & 0.123 \\
Integrated Gradients & 0.071 & 0\% & 0.074 \\
LIME & 0.068 & 0\% & 0.049 \\
\bottomrule
\end{tabular}
\caption{Encoder results on BERT-base-uncased. $^\dagger$417/500 after filtering.}
\label{tab:encoder_results}
\end{table}

\section{Effect Sizes and Anti-Faithfulness Details}
\label{app:effect_sizes}

Tables~\ref{tab:effect_sizes}--\ref{tab:bootstrap_ci} provide effect sizes, anti-faithful configurations, concrete anti-faithful examples, and bootstrap confidence intervals.

\begin{table}[H]
\centering
\footnotesize
\setlength{\tabcolsep}{3pt}
\begin{tabular}{@{}lccc@{}}
\toprule
\textbf{Configuration} & \textbf{WR} & \textbf{$d_n$} & \textbf{Interp.} \\
\midrule
Llama 3.1 e-SNLI Grad & 97.2\% & 2.50 & Ext.\ large \\
Qwen IMDB Attn & 94.9\% & 1.96 & V.\ large \\
Qwen IMDB Grad & 91.4\% & 1.84 & Large \\
Llama 3.2 e-SNLI Attn & 86.4\% & 3.77 & V.\ large \\
Qwen DE Attn & 82.7\% & 1.40 & Large \\
Llama 3.1 FR Attn & 80.8\% & 1.26 & Large \\
\midrule
GPT-2 FR Attn & 15.8\% & -2.08 & Anti \\
GPT-2 FR Grad & 14.8\% & -2.36 & Anti \\
GPT-2 e-SNLI Grad & 29.5\% & -0.72 & Anti \\
DeepSeek AG Grad & 39.1\% & -0.53 & Anti \\
\bottomrule
\end{tabular}
\caption{Standardized effect sizes ($d_n$). $d_n > 0.8$ = large. Anti = anti-faithful.}
\label{tab:effect_sizes}
\end{table}

\begin{table}[H]
\centering
\footnotesize
\setlength{\tabcolsep}{3pt}
\begin{tabular}{@{}llcc@{}}
\hline
\textbf{Model} & \textbf{Config} & \textbf{WR} & \textbf{$d_n$} \\
\hline
GPT-2 & FR Attn/Grad & 15--16\% & -2.1 \\
GPT-2 & e-SNLI Grad & 29.5\% & -0.72 \\
GPT-2 & DE Grad & 34.2\% & -0.39 \\
DeepSeek & SST-2 Grad & 40.5\% & -0.29 \\
DeepSeek & AG Grad & 39.1\% & -0.53 \\
Llama 3.2 & SST-2 Grad & 42.4\% & -0.44 \\
Llama 3.1 & SST-2 Grad & 46.0\% & -0.15 \\
Llama 3.1 & AG Grad & 42.2\% & -0.40 \\
\hline
\end{tabular}
\caption{Anti-faithful configurations (WR $<$ 50\%). Negative $d_n$ = worse than random.}
\label{tab:anti_faithful_configs}
\end{table}

\begin{table}[H]
\centering
\footnotesize
\setlength{\tabcolsep}{2pt}
\begin{tabular}{@{}p{3.2cm}ccp{2.8cm}@{}}
\hline
\textbf{Text} & \textbf{WR} & \textbf{$d_n$} & \textbf{Pattern} \\
\hline
``gorgeous, witty, seductive'' & 0\% & -2.3 & Selects \textit{a}; ignores adjectives \\
``tender, heartfelt drama'' & 0\% & -1.1 & Selects \textit{a} \\
``fast, funny, enjoyable'' & 0\% & -2.3 & Selects \textit{a} \\
``high comedy, poignance'' & 0\% & -2.8 & Selects \textit{uses} \\
\hline
\end{tabular}
\caption{Anti-faithful examples (Llama 3.1-8B/SST-2). Gradient selects initial function words, ignoring sentiment.}
\label{tab:anti_faith_examples}
\end{table}

\begin{table}[H]
\centering
\small
\begin{tabular}{lcc}
\toprule
\textbf{Configuration} & \textbf{Win Rate} & \textbf{95\% CI} \\
\midrule
Llama 3.1 e-SNLI Grad & 97.2\% & [95.4, 99.0] \\
Qwen IMDB Attn & 94.9\% & [92.1, 97.7] \\
Llama 3.2 e-SNLI Attn & 86.4\% & [82.1, 90.7] \\
\midrule
GPT-2 DE Grad & 34.2\% & [28.7, 39.7] \\
GPT-2 FR Attn & 15.8\% & [11.2, 20.4] \\
\bottomrule
\end{tabular}
\caption{Bootstrap 95\% CIs for uncertainty quantification. Formal significance is assessed via BH-corrected randomization-test $p$-values.}
\label{tab:bootstrap_ci}
\end{table}

\section{Faithfulness-Plausibility Correlation}
\label{app:alignment}

Table~\ref{tab:alignment_appendix} reports IoU-faithfulness correlations across three models on e-SNLI.

\begin{table}[H]
\centering
\small
\begin{tabular}{llccc}
\toprule
\textbf{Model} & \textbf{Method} & \textbf{r} & \textbf{p} & \textbf{N} \\
\midrule
GPT-2 (1.5B) & Attention & 0.016 & 0.73 & 462 \\
             & Gradient  & -0.018 & 0.69 & 493 \\
\midrule
DeepSeek-7B  & Attention & 0.033 & 0.53 & 370 \\
             & Gradient  & 0.019 & 0.68 & 487 \\
\midrule
Mistral-7B   & Attention & 0.016 & 0.77 & 351 \\
             & Gradient  & 0.012 & 0.79 & 485 \\
\bottomrule
\end{tabular}
\caption{IoU-Faithfulness correlation across three models. No model shows significant correlation ($|r| < 0.04$, all $p > 0.5$).}
\label{tab:alignment_appendix}
\end{table}

\section{Retrieval Infill Detailed Results}
\label{app:retrieval}

\begin{figure}[t]
\centering
\resizebox{\columnwidth}{!}{%
\begin{tikzpicture}
\begin{axis}[
    width=\columnwidth,
    height=7.5cm,
    xmin=38, xmax=92,
    ymin=0.3, ymax=6.7,
    xlabel={Win Rate (\%)},
    xlabel style={font=\small},
    xtick={40,50,60,70,80,90},
    xticklabel style={font=\footnotesize},
    ytick={1,2,3,4,5,6},
    yticklabels={
        {GPT-2 / SST-2},
        {DeepSeek / SST-2},
        {Mistral / AG},
        {Qwen / SST-2},
        {Qwen / e-SNLI},
        {Llama-3.2 / e-SNLI}
    },
    yticklabel style={font=\footnotesize},
    ytick align=outside,
    grid=major,
    grid style={line width=0.2pt, draw=black!12},
    major grid style={line width=0.3pt, draw=black!15},
    axis line style={draw=black!40},
    tick style={draw=black!40},
    clip=true,
    legend style={
        at={(0.98,0.02)},
        anchor=south east,
        font=\footnotesize,
        draw=black!30,
        fill=white,
        fill opacity=0.9,
        text opacity=1,
        row sep=1pt,
        inner sep=3pt,
        legend cell align=left,
    },
]

\draw[dashed, black!40, line width=0.7pt] (axis cs:50,0.3) -- (axis cs:50,6.7);
\node[font=\tiny, text=black!50, anchor=south, rotate=90] at (axis cs:50,6.7) {random baseline};

\draw[black!25, line width=0.8pt] (axis cs:42.6,6) -- (axis cs:86.4,6);
\draw[black!25, line width=0.8pt] (axis cs:77.3,5) -- (axis cs:77.5,5);
\draw[black!25, line width=0.8pt] (axis cs:59.6,4) -- (axis cs:68.4,4);
\draw[black!25, line width=0.8pt] (axis cs:42.7,3) -- (axis cs:51.7,3);
\draw[black!25, line width=0.8pt] (axis cs:53.9,2) -- (axis cs:62.2,2);
\draw[black!25, line width=0.8pt] (axis cs:52.4,1) -- (axis cs:60.6,1);

\addplot[
    only marks,
    mark=*,
    mark size=2.5pt,
    color=red!75!black,
    fill=red!75!black,
] coordinates {
    (86.4, 6)
    (77.3, 5)
    (68.4, 4)
    (51.7, 3)
    (62.2, 2)
    (60.6, 1)
};

\addplot[
    only marks,
    mark=*,
    mark size=2.5pt,
    color=blue!70!black,
    fill=blue!70!black,
] coordinates {
    (42.6, 6)
    (77.5, 5)
    (59.6, 4)
    (42.7, 3)
    (53.9, 2)
    (52.4, 1)
};

\legend{Delete, Retrieval Infill}

\node[font=\tiny, text=red!60!black, anchor=west] at (axis cs:87,6) {+44};
\node[font=\tiny, text=black!50, anchor=west] at (axis cs:78,5) {$\approx$0};
\node[font=\tiny, text=red!60!black, anchor=west] at (axis cs:69,4) {+9};
\node[font=\tiny, text=red!60!black, anchor=west] at (axis cs:52.2,3) {+9};
\node[font=\tiny, text=red!60!black, anchor=west] at (axis cs:63,2) {+8};
\node[font=\tiny, text=red!60!black, anchor=west] at (axis cs:61,1) {+8};

\end{axis}
\end{tikzpicture}%
}
\caption{
    Paired comparison of attention win rates under Delete vs.\ Retrieval Infill (main models).
    Delete ({\color{red!75!black}red}) exceeds Retrieval Infill on most short-text configurations by 8--44\,pp,
    but operators \emph{agree} on Qwen/e-SNLI ($\approx$77\%), suggesting genuine faithfulness there.
}
\label{fig:operator_comparison}
\end{figure}
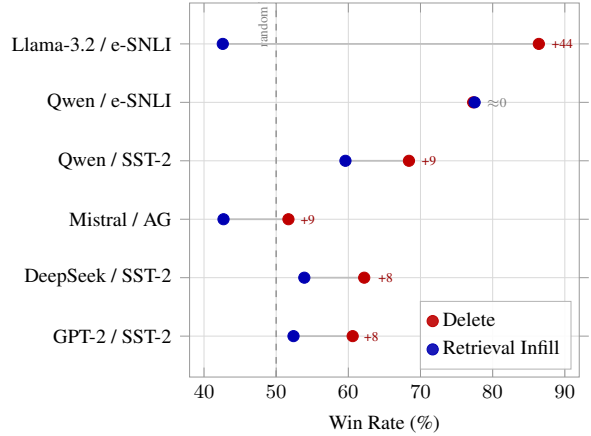

Table~\ref{tab:retrieval_comparison_appendix} and Figure~\ref{fig:operator_comparison} provide representative operator comparisons.

\begin{table}[H]
\centering
\small
\begin{tabular}{l|cc|cc}
\toprule
& \multicolumn{2}{c|}{\textbf{Delete}} & \multicolumn{2}{c}{\textbf{Retrieval}} \\
\textbf{Configuration} & WR & Tax & WR & Tax \\
\midrule
Qwen / SST-2 & \textbf{68.4} & F & 59.6 & WF \\
Qwen / e-SNLI & \textbf{77.3} & F & 77.5 & F \\
Llama-3.2 / e-SNLI & \textbf{86.4} & F & 42.6 & AF \\
Mistral / AG News & 51.7 & NR & 42.7 & AF \\
DeepSeek / SST-2 & \textbf{62.2} & F & 53.9 & NR \\
GPT-2 / SST-2 & \textbf{60.6} & F & 52.4 & NR \\
\bottomrule
\end{tabular}
\caption{Attention win rates for Delete vs.\ Retrieval Infill ($k=0.2$). F=Faithful ($>$60\%), WF=Weakly faithful (55--60\%), NR=Near-random (50--55\%), AF=Anti-faithful ($<$50\%).}
\label{tab:retrieval_comparison_appendix}
\end{table}


\section{Relocated English Results Table}
\label{app:english_results}

Table~\ref{tab:english_results_appendix} provides the full English attention win rates underlying the top panels of Figure~\ref{fig:english_heatmap}. Table~\ref{tab:guidelines_appendix} summarizes practical recommendations.

\begin{table}[H]
\centering
\small
\begin{tabular}{lcccc}
\toprule
\textbf{Model} & \textbf{SST-2} & \textbf{IMDB} & \textbf{e-SNLI} & \textbf{AG News} \\
\midrule
\multicolumn{5}{c}{\textit{Deletion --- Attention Win Rate (\%)}} \\
\midrule
GPT-2 & 60.6 & 44.0 & 64.0 & \textbf{70.8} \\
Llama 3.2-3B & 53.2 & 71.3 & \textbf{86.4} & 56.0 \\
Llama 3.1-8B & 53.2 & 52.5 & 85.2 & 47.5 \\
Qwen 2.5-7B & \textbf{68.4} & \textbf{94.9} & 77.3 & 62.2 \\
Mistral 7B & 57.5 & 83.8 & 71.1 & 51.7 \\
DeepSeek 7B & 62.2 & 84.6 & 63.7 & 49.4 \\
LFM2-2.6B & 45.4 & 50.3 & 66.6 & 57.0 \\
\midrule
\multicolumn{5}{c}{\textit{Retrieval --- Attention Win Rate (\%)}} \\
\midrule
GPT-2 & 52.4 & 38.3 & 68.2 & \textbf{73.2} \\
Llama 3.2-3B & 52.7 & \textbf{91.8} & 42.6 & 56.9 \\
Llama 3.1-8B & 51.5 & 75.3 & 74.0 & 49.4 \\
Qwen 2.5-7B & 59.6 & \textbf{96.3} & \textbf{77.5} & 59.8 \\
Mistral 7B & 56.1 & 89.8 & 67.5 & 42.7 \\
DeepSeek 7B & 53.9 & 80.5 & 65.5 & 46.7 \\
LFM2-2.6B & 45.5 & 49.4 & 65.5 & 51.1 \\
\bottomrule
\end{tabular}
\caption{English attention win rates under both operators. Bold = best per column/operator.}
\label{tab:english_results_appendix}
\end{table}

\begin{table}[H]
\centering
\resizebox{\columnwidth}{!}{
\begin{tabular}{ll}
\hline
\textbf{Scenario} & \textbf{Recommendation} \\
\hline
Short text, Sentiment & Use Attention \\
Long text & Either method works \\
NLI & Use Attention \\
Topic Classification & Use Attention (Gradient for Mistral) \\
\hline
French & Llama 3.1 Attn (81\%) or LFM2 Grad (73\%) \\
German & Qwen Attn (83\%) -- others struggle \\
Chinese & LFM2 Grad (69\%) or Llama 3.1 Grad (66\%) \\
Hindi & GPT-2 (65--66\%) works surprisingly well \\
Turkish & Qwen Attn (79\%) or Mistral Grad (72\%) \\
Arabic & Mistral Attn (73\%) -- avoid Llama-3.2 (37\%) \\
\hline
\end{tabular}
}
\caption{Practical attribution guidelines based on \ice{} evaluation.}
\label{tab:guidelines_appendix}
\end{table}

\section{Preliminary: Extension to Chain-of-Thought}
\label{app:cot}

This appendix presents \textbf{preliminary} evidence that ICE's operator-aware methodology generalizes from token-level attributions to step-level chain-of-thought (CoT) evaluation. The core insight---comparing against random baselines under multiple operators---transfers when the unit of analysis changes from tokens to reasoning steps. A comprehensive evaluation across frontier models is the subject of separate work.

\subsection{Step-Level Operators}

Given a CoT response $C = [s_1, s_2, \ldots, s_n]$ with final answer $a$, we define step-level analogues of ICE's operators:
\begin{itemize}
    \item \textbf{Step-Delete}: Remove step $s_i$, concatenate remaining steps, query the model for a new answer.
    \item \textbf{Step-Retain}: Present only a subset of $k$ steps, query for an answer.
\end{itemize}
Aggregating over random subsets of size $\lfloor k \cdot n \rfloor$ with $P$ permutations yields step-level necessity and sufficiency win rates, paralleling ICE's token-level NSR decomposition. We combine these into step-level reasoning capacity $= \text{Nec}_{k=0.2} \times (1 - \text{Suf}_{k=0.2})$. ICE's statistical testing (randomization tests, bootstrap CIs, BH correction) applies identically.

\subsection{Operator Sensitivity at Step Level}

A key finding of our token-level analysis is that faithfulness estimates are operator-dependent. We verify this persists at the step level using 3 open-weight models on GSM8K ($N=50$):

\begin{table}[h]
\centering
\small
\begin{tabular}{lcccc}
\toprule
\textbf{Model} & \multicolumn{2}{c}{\textbf{Token-level}} & \multicolumn{2}{c}{\textbf{Step-level}} \\
\cmidrule(lr){2-3} \cmidrule(lr){4-5}
& Nec & Suf & Nec & Suf \\
\midrule
Qwen3-0.6B & 0.48 & 0.61 & 0.08 & 0.04 \\
Qwen3-8B & 0.77 & 0.50 & 0.85 & 0.12 \\
DeepSeek-R1-7B & 0.13 & 0.57 & 0.80 & 0.17 \\
\bottomrule
\end{tabular}
\caption{Token-level vs.\ step-level faithfulness on GSM8K. DeepSeek-R1-7B appears unfaithful at the token level (Nec$=0.13$) but faithful at the step level (Nec$=0.80$), demonstrating that evaluation granularity---tokens vs.\ steps---is itself an operator choice.}
\label{tab:step_vs_token}
\end{table}

\subsection{Black-Box Frontier Model Evaluation}

Unlike token-level ICE, step-level evaluation operates through standard chat completions: remove a sentence from the prompt and re-query. This enables evaluation of closed-source frontier models. We demonstrate on 3 models ($N=500$):

\begin{table}[h]
\centering
\small
\begin{tabular}{lcccc}
\toprule
& \multicolumn{2}{c}{\textbf{GSM8K}} & \multicolumn{2}{c}{\textbf{SST-2}} \\
\cmidrule(lr){2-3} \cmidrule(lr){4-5}
\textbf{Model} & SRC & Acc & SRC & Acc \\
\midrule
Model A (RL-trained) & 0.86 & 93\% & 0.87 & 97\% \\
Model B (RL-trained) & 0.86 & 84\% & 0.22 & 87\% \\
Model C (instruction) & 0.01 & 95\% & 0.00 & 97\% \\
\bottomrule
\end{tabular}
\caption{Step-level reasoning capacity (SRC $= \text{Nec}_{0.2} \times (1 - \text{Suf}_{0.2})$) for frontier models. High accuracy does not imply faithful reasoning: Model C matches Model A's accuracy but has SRC$\approx$0 (decorative reasoning).}
\label{tab:frontier_cot}
\end{table}

The results suggest a three-mode taxonomy: genuine (SRC $> 0.4$), scaffolding ($0.15$--$0.4$), and decorative ($< 0.15$) reasoning. The cost is ${\sim}$\$1--2 per model per task---enabling routine deployment auditing.

\paragraph{Reproducibility.} Code, results, pre-trained extractors, cached win rates, and reproduction scripts are provided in the supplementary material.

\end{document}